\documentclass[10pt,twocolumn,letterpaper]{article}

\usepackage{cvpr}              % To produce the CAMERA-READY version
\usepackage{float}
\usepackage{abstract}
\usepackage[ruled,vlined,linesnumbered]{algorithm2e}
\usepackage{multirow}
\usepackage{graphics}
\usepackage{pifont}
\usepackage{footnote}
\usepackage[normalem]{ulem}
\usepackage[misc,geometry]{ifsym}
\definecolor{cvprblue}{rgb}{0.21,0.49,0.74}
\usepackage[pagebackref,breaklinks,colorlinks,allcolors=cvprblue]{hyperref}
\newcommand{\cmark}{\ding{51}}%
\newcommand{\xmark}{\ding{55}}

\def\paperID{8537} % *** Enter the Paper ID here
\def\confName{CVPR}
\def\confYear{2025}

\title{\textit{BADGR}: Bundle Adjustment Diffusion Conditioned by GRadients for Wide-Baseline Floor Plan Reconstruction}
\author{Yuguang Li$^{1,2}$\textsuperscript{\Letter}\thanks{Work done at University of Washington.}\hspace{0.6cm} 
Ivaylo Boyadzhiev\thanks{Work done as an independent researcher.}\hspace{0.6cm} 
Zixuan Liu$^1$\hspace{0.6cm} 
Linda Shapiro$^{1}$\thanks{Equal contribution.}\hspace{0.6cm} 
Alex Colburn$^{1}\footnotemark[3]$ 
\\
$^1$University of Washington \hspace{0.6cm} $^2$Zillow Group \\
\vspace{-2pt}
}

\begin{document}
% \raggedbottom

\twocolumn[{%
\maketitle
\renewcommand\twocolumn[1][]{#1}%
   \vspace{-12mm} 
    \centering
    \includegraphics[width=1\linewidth]{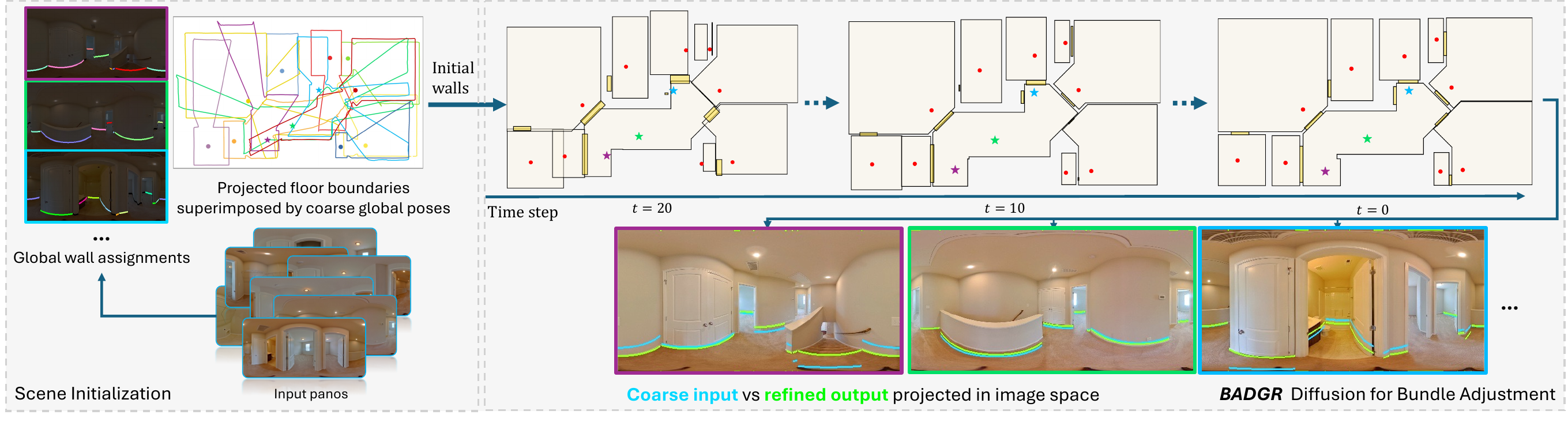}
    \vspace{-10mm} 
    \captionsetup{hypcap=false}  % Disable hypcap
    \captionof{figure}{\footnotesize 
    Overview of \textit{BADGR}, a diffusion-based bundle adjustment (\textit{BA}) model for generating precise, view-consistent camera poses and floor plan layouts. \textit{BADGR} uses per-image floor boundaries and image column-to-wall assignments (upper left) as coarse input, refining poses and layouts through a gradient-conditioned denoising process (upper right). The bottom right shows view consistency by projecting the output layouts with the estimated poses.}
    \label{fig:overview}
    \vspace{7pt} 
  % \vspace{5mm} 
 }]

\pagestyle{empty}
\thispagestyle{empty}

\saythanks
% \maketitle

% \begin{figure*}[ht]
% \begin{center}
%     \includegraphics[width=0.95\linewidth]{figs/overview_v2.pdf}
%     \caption{Floor plan and camera pose estimation with \textit{BADGR}. Designed to be similar to \textbf{PoseDiffusion} first figure. \yuguang{Should it be right under the title?}}
%     \label{fig:overview}
% \end{center}
% \end{figure*} 

\vspace{-4pt}
\begin{abstract}
\vspace{-4pt}

Reconstructing precise camera poses and floor plan layouts from wide-baseline RGB panoramas is a difficult and unsolved problem. We introduce \textit{BADGR}, a novel diffusion model that jointly performs reconstruction and bundle adjustment (BA) to refine poses and layouts from a coarse state, using 1D floor boundary predictions from dozens of sparsely captured images. Unlike guided diffusion models, \textit{BADGR} is conditioned on dense per-column outputs from a single-step \textit{Levenberg Marquardt} (\textit{LM}) optimizer and is trained to predict camera and wall positions, while minimizing reprojection errors for view consistency. The objective of layout generation from denoising diffusion process complements \textit{BA} optimization by providing additional learned layout-structural constraints on top of the co-visible features across images. These constraints help \textit{BADGR} make plausible guesses about spatial relationships, which constrain the pose graph, such as wall adjacency and collinearity, while also learning to mitigate errors from dense boundary observations using global context. \textit{BADGR} trains exclusively on 2D floor plans, simplifying data acquisition, enabling robust augmentation, and supporting a variety of input densities. Our experiments validate our method, which significantly outperforms the state-of-the-art pose and floor plan layout reconstruction with different input densities. Visit project website at: \href{https://badgr-diffusion.github.io}{https://badgr-diffusion.github.io}.

\end{abstract}
    
\section{Introduction}
\label{sec:intro}

Reconstructing floor plan layouts and camera poses has become an important task with many applications such as virtual touring, interior design, and autonomous navigation. High spatial accuracy in both objectives is essential for high-fidelity downstream tasks, such as cross-view scene editing and dense reconstruction. Existing solutions for image-based layout reconstruction are either coarse, limited to a single room, or, while more accurate, require either densely captured image inputs or sparser capture with RGB-D cameras, which can be costly in terms of equipment, data bandwidth and capture efforts.

This work aims to accurately reconstruct camera extrinsics and floor plan layouts from sparsely captured $360^\circ$ panoramas without prior pose information \cite{cruz2021zillow}. Specifically, our goals are to: (1) reconstruct the floor plan as a unique set of closed-loop polygons defining rooms and doors \cite{yue2023roomformer, cruz2021zillow}; (2) estimate each camera pose for view-consistency \cite{furukawa2015_multiview_tutorial}; (3) accommodate diverse capture densities, down to one image per room; and (4) ensure that generated floor plans remain plausible within the natural distribution of training data, even when certain walls are occluded.
 
Accurate reconstruction of camera poses and spatial layouts in wide-baseline indoor environments is challenged by limited co-visibility and sparse features. This process demands not only view-consistency but also layout-structural constraints, such as \textit{Manhattan} or \textit{Atlanta} frameworks \cite{schindler2004atlanta}, wall thickness, collinearity, and prior knowledge of room layouts \cite{zheng2020structured3d, cruz2021zillow, lambert2022salve}. We propose \textit{BADGR}, a conditional denoising diffusion probabilistic model (\textit{DDPM}) trained to reconstruct and bundle adjust camera poses and layouts. \textit{BADGR} employs a planar bundle adjustment (\textit{BA}) module, to provide geometric guidance for conditioning the \textit{DDPM} to maximize view-consistency from angle-constrained layouts and poses. The \textit{DDPM} is also trained for the floor plan generation task. Combined with a reprojection loss, \textit{BADGR} performs bundle adjustment through posterior sampling, with learned layout-structural constraints and the ability to handle noise from input features. The generative ability of the \textit{DDPM} allows \textit{BADGR} to predict plausible shapes of occluded layout sections based on training data. The \textit{non-Markovian} inference process, i.e., predicting $x_{t-1}$ from $x_{t}$ by predicting $x_{0}$, allows \textit{BADGR} to combine a score-based generative model with a nonlinear optimization process, without specifying step size during training. \textit{BADGR} differs from guided-diffusion style models \cite{dhariwal2021_guided_diffusion, chung2023_blind_inverse_operator, chung2023_3d_inverse, wang2023posediffusion}, where the gradients from differentiable objective functions are used only during inference to guide a pre-trained diffusion model,
leading to issues like slow convergence and deviation from data manifold.  
Our experiments show that \textit{BADGR} is more accurate in performing \textit{BA} tasks compared to guided-diffusion style models.

The contributions of this work are: 1) \textit{BADGR} is the first learning-based approach to jointly refine deformable room layouts from polygon-based floor plans and camera poses from sparsely captured RGB panoramas, guided by visually-derived features; 
2) \textit{BADGR} contains a novel approach to train a diffusion model for both nonlinear optimization (i.e., \textit{BA}) and generation tasks (i.e., floor plan generation), allowing it to reconstruct poses and layouts from visual inputs and make reasonable guesses from learned layout-structural constraints; 3) \textit{BADGR} obtains state-of-the-art accuracy in wide-baseline camera pose estimation and layout estimation with multi-view 360$^\circ$ panoramas.
\section{Related Work}
\label{sec:related}

Our work bridges floor plan reconstruction and wide-baseline pose estimation, jointly reasoning over layout constraints and cross-view consistency.

\noindent \textbf{Image-based Floor Plan Reconstruction} involves estimating camera poses and creating a unique set of layout polygons. Research has shown good accuracy in producing floor plans in such formats from registered RGB-D point cloud scans \cite{chen2019floorsp, Chen19iccv_FloorSPInverseCAD, stekovic2021montefloor, chen2022heat, yue2023roomformer, chen2024polydiffuse}. Generative models \cite{Nauata20eccv_HouseGAN, Nauata21cvpr_HouseGAN++, shabani2023housediffusion} also demonstrate deep learning’s ability to model layout constraints, enabling floor plan generation from abstract inputs, like bubble diagrams \cite{Nauata21cvpr_HouseGAN++}. Over the years, single-view room layout estimation has been extensively studied \cite{zou2018layoutnet, sun2019horizonnet, pintore2020atlantanet, sun2021hohonet, wang2021led2, jiang2022lgtnet}. However, noisy or unknown poses remain a key challenge for multi-view, wide-baseline layout reconstruction, as it requires 1) generating a single fused layout, 2) high accuracy in both camera poses and layouts for consistent views. 
Prior approaches attempted to regress poses and layouts from panorama pairs \cite{wang2022psmnet, su2023gpr, lambert2022salve}, yet often lack view-consistency
without joint optimization \cite{wang2022psmnet}.

\noindent \textbf{Wide-Baseline Pose Estimation} Analysis has shown \cite{lambert2022salve, hutchcroft2022covispose} that wide-baseline indoor pose estimation is challenging to traditional \textit{Structure from Motion} (\textit{SfM}), as images often contain featureless regions, repetitive textures, or narrow passageways, which limit co-visibility and cause drastic appearance changes across images \cite{cruz2021zillow}. Semantic constraints can help \textit{SfM} establish anchors and loop-closures, but existing approaches often rely on heuristics and lack understanding of structural layouts  \cite{Cohen15iccv_StitchingDisconnected, Cohen16eccv_IndoorOutdoorAlignment}. \cite{shabani2021extreme, lambert2022salve} explored wide-baseline reconstruction as discrete problems, piecing rooms together like puzzles \cite{hossieni2024puzzlefusion} and using vanishing points \cite{zhang2014panocontext} to improve camera rotation estimates. However, these approaches can produce coarse poses and layouts with missing rooms from a larger floor plan. Recently, direct pose regression achieved good retrieval accuracy in solving coarse poses from a pair of \cite{chen2021_directionnet, hutchcroft2022covispose, tu2024panopose} or up-to-5 \cite{nejatishahidin2023graph} panorama inputs, along with predictions of dense correspondences among image columns and dense room layouts \cite{hutchcroft2022covispose, su2023gpr}. A global pose graph can be later formed by merging local poses \cite{nejatishahidin2023graph, lambert2022salve}, but this often lacks sufficient accuracy for good view-consistency. Furthermore, structural-layout constraints are frequently violated when superimposing projected single-view layouts on estimated poses \cite{wang2022psmnet}, as shown in the starting floor plan of Figure \ref{fig:overview}.

\begin{figure*}[ht]
\begin{center}
    \vspace{-8pt}
    \centering
    \includegraphics[width=0.95\linewidth]{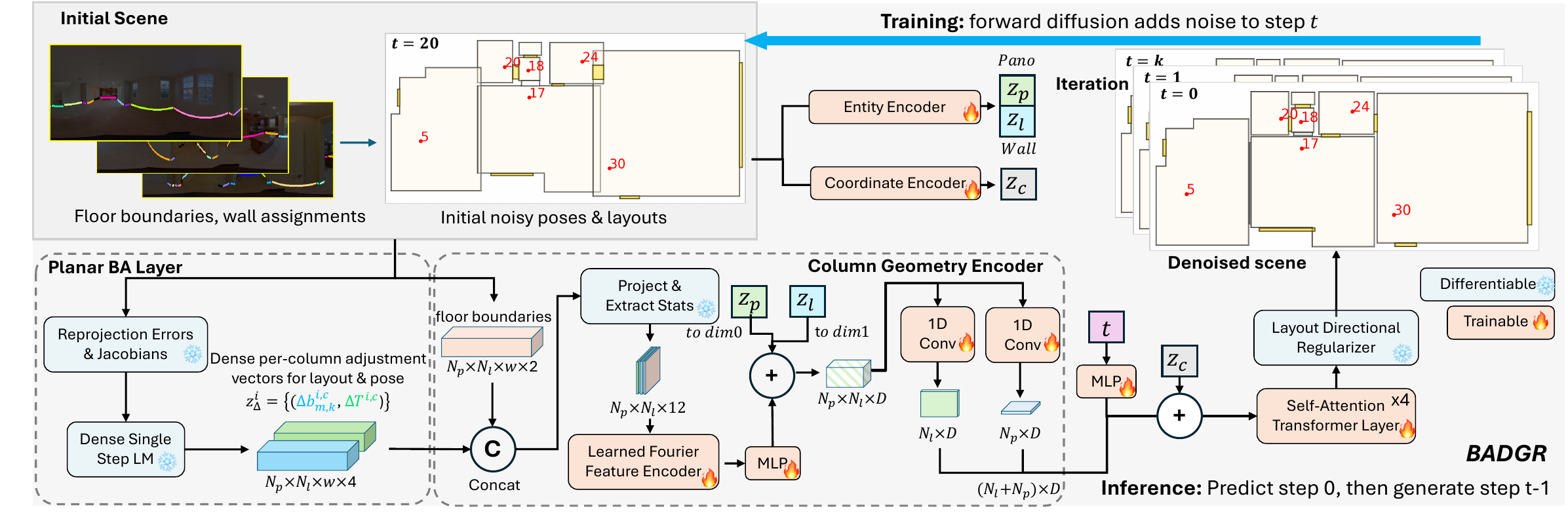}
    \vspace{-10pt}
    \caption{Architecture of \textit{BADGR}. The forward process takes a ground truth scene, i.e. layouts and poses, adds noise to sample step t. The inference process uses a transformer, conditioned on dense per-column adjustments generated by the planar \textit{BA} layer and compressed by the Column Geometry Encoder.}
    \label{fig:Architecture}
\end{center}
\vspace{-26pt}
\end{figure*}

\noindent \textbf{Learned Optimization} In traditional \textit{SfM}, robust optimization often integrates sensor data and physical constraints iteratively to manage uncertainties in measurements and features. Recent data-driven approaches improved "front-end" feature extraction and matching \cite{sarlin2020superglue, pautrat2023gluestick}, adapting to noise and variability more effectively than classical methods. Differentiable optimizer, like differentiable \textit{LM} \cite{pineda2022theseus, tang2018_banet}, have been used to train feature extraction with uncertainty prediction \cite{muhle2023learning}. Uncertainty is also modeled from the parameter posterior distributions given initial measurements. \textit{G3R} \cite{chen2025g3r} iteratively optimizes 3D Gaussians using a gradient-conditioned 3D \textit{U-Net}. \textit{PoseDiffusion} \cite{wang2023posediffusion}, \textit{PhysDiff} \cite{yuan2023physdiff} apply posterior sampling of pre-trained diffusion models for improved reconstruction. However, posterior-guided sampling can run into conflicts between guidance and diffusion flow function, which affects convergence \cite{chung2022_manifold}. \textit{Non-Markovian} \textit{DDPMs}, such as denoising diffusion implicit models (\textit{DDIM}) \cite{song2020_ddim}, enable conditional diffusion training without explicit transitions from t to t-1, facilitating integration with nonlinear optimization independent of step size.

\vspace{-10pt}
\section{Problem Statement}
\label{sec:statement}
\vspace{-2pt}

\textbf{Objectives}:
Given a set of sparsely captured $360^\circ$ indoor RGB panorama images $\{P^i\}$ in equirectangular projections without pose information, the overall pipeline 
aims to estimate 3 degrees-of-freedom (\textit{DoF}) camera poses $\{E^i, E^i\in SE(2)\}$, i.e. $xy$ position, horizontal rotation angle, and 2D floor plans represented as a set of closed-loop polygons $\{V_{m}, V_m=( v_{m,1}, v_{m,2}, ..., v_{m,k})\}$ for rooms and doors in a single global coordinate system, where $m$, $k$ is room id and corner id, and $v_{m,k}$ is a 2D $xy$ vertex coordinate. Our main focus, the proposed \textit{BA}-based refinement component \textit{BADGR}, aims to denoise camera $xy$ positions $\mathcal{T}^i$ and angular constrained layouts $\{V_{m}\}$ by moving walls along their normal directions. \textit{BADGR} optimizes for view-consistency on re-projected floor boundaries, assuming each image column contributes to the positional adjustment of a single wall and a camera, where the column-to-wall relations have been pre-assigned. 
\textit{BADGR} also learns to maintain the layout-structural constraints of floor plans from training data. 

\noindent\textbf{Assumptions}: We assume that all images are straightened \cite{zhang2014panocontext} and can be connected to a single pose graph with co-visibility \cite{cruz2021zillow} to nearest neighbor greater than 10\%. Relative camera heights are known across the floor. During \textit{BADGR} optimization, wall angles are assumed to be fixed. The number of walls and wall connectivity are given; they can either come from merging the single-view layout estimation, post-processed under the \textit{Atlanta World} assumption \cite{jiang2022lgtnet, pintore2020atlantanet}, or via a quick human-in-the-loop process, see Section \ref{sec:init}. Our floor plans follow the \textit{Atlanta World} assumption \cite{schindler2004atlanta} and have no curved walls. 

\vspace{-4pt}
\section{Pipeline Overview}

We designed a coarse-to-fine pipeline, as shown in Figure \ref{fig:overview}, that initializes a scene with estimated camera poses and room layouts, as closed-loop polygons, combining per-view floor boundary predictions and image column-to-wall assignments, similar to semantic column matches across multiple images \cite{wang2021led2, agarwal2011_building_rome}. This coarse initialization is subsequently refined with \textit{BADGR}, which focuses on accurate reconstruction. Starting from the coarse scene, rotations from estimated poses and walls are corrected via vanishing point snapping \cite{zhang2014panocontext}. During \textit{BADGR} refinement, the scene is angularly constrained: \textit{BADGR} optimizes only the 2-\textit{DoF} camera \textit{xy} position \(\mathcal{T}^i\) for each pose \(E^i\) and the 1-\textit{DoF} line translation \(b_{m,k}\) for each wall \(l_{m,k}\). While various algorithms could initialize the coarse floor plan for \textit{BADGR}, we outline a specific pipeline in Section \ref{sec:init} using practical algorithms for coarse scene setup. The scene representation is discussed in Section \ref{sec:scene_representation}, with details of \textit{BADGR}, the proposed learned \textit{BA} diffusion model, in Section \ref{sec:badgr_diffusion}.

\section{Coarse Scene Initialization}
\label{sec:init}
During inference, the scene is initialized with a slightly modified \textit{CovisPose} model \cite{hutchcroft2022covispose}. \textit{CovisPose} is run on each pair of panoramas on the same floor to predict 1) relative camera pose $\Tilde{E}^{(i,j)}\in SE(2)$, 2) floor boundaries $\{\Tilde{\mathcal{B}}^i \}$, and 3) cross-view co-visibility, angular correspondences $\{\Tilde{\alpha}^{i,j} \}$, $\{\Tilde{\varphi}^{i,j} \}$. Additionally, we perform per-column binary classification to identify room corners $\{\Tilde{\mathcal{V}}^i \}$. The model is trained on the \textit{ZInD} dataset \cite{cruz2021zillow} with the same image pairs as \cite{hutchcroft2022covispose} with an additional corner loss function similar to that of \cite{sun2019horizonnet}. Pose pairs of co-visibility score greater than 0.1 are selected to create a minimal spanning tree of the pose graph, similar to \cite{nejatishahidin2023graph}.  $\Tilde{E}^{i,j}$ are corrected through axis alignment with a 45° interval using predicted vanishing angles \cite{zhang2014panocontext} prior to computing global poses $\Tilde{E}^i$.  

The per-pano floor boundaries $\{\hat{\mathcal{B}}^i \}$ are further refined and aggregated into uniquely identifiable set of global walls $l_{m,k}$, shared across ${P^i}$ via an automatic process using room corners $\{\Tilde{\mathcal{V}}^i \}$ . Finally, an annotator uses an interactive application to provide global wall connectivity, and add missing room corners with their rough initial positions.  The number of room corners and wall orientations are static input to \textit{BADGR}. More details are provided in Supplementary.

\section{Scene Representation for \textit{\textbf{BA}} Optimization}
\label{sec:scene_representation}

Our proposed representation aims to uniquely define walls and cameras while assigning image columns to global walls, allowing \textit{BA} to perform cross-view reprojections at any scene state. Each column links either to a specific wall or remains unassigned, simplifying reprojections by avoiding wall occlusion handling during floor boundary rendering. The global scene comprises room layouts \(\{V_m\}\), doors as polygons with at max $N_l$ walls, camera extrinsics \(\{E^i\}\) of $N_p$ panoramas, per-panorama floor boundaries \(\{\hat{\mathcal{B}}^{i, c} \}\) for columns \(\{\mathcal{C}^{i, c}\}\) with image width $w$, and a column-to-wall semantic assignment represented as a one-hot 3D array of shape \(N_p \times w \times N_l\) for mapping \(\{\mathcal{M}: \mathcal{C}^{i, c} \rightarrow l_{m,k}\}\). 
The layout walls \(\{V_m\}\) are represented as line segments \(\{l_{m,k} \mid l_{m,k} = (v_{m,k}, v_{m,k+1})\}\) with each line represented in the Hesse  normal form \cite{hesse_normal_form}
allowing us to easily work with rotations and offsets \((\overrightarrow{v_{m,k}}, b_{m,k})\) from the origin.
\textit{BADGR} optimizes layout vertices \(V_m\) and camera positions \(\mathcal{T}^i\), with these parameters normalized within \([-1, 1]\) as a continuous 2D coordinate array. Constant scene parameters include camera and wall rotation vectors \(\mathcal{R}_i\), \(\overrightarrow{v_{m,k}}\), camera height \(z^i\) and column-to-wall assignments \(\{\mathcal{M}\}\). Details on scene initialization are covered in Section \ref{sec:init}.

\section{Bundle Adjustment Diffusion}
\label{sec:badgr_diffusion}
\begin{figure}
\begin{center}
    \includegraphics[width=1.0\linewidth]{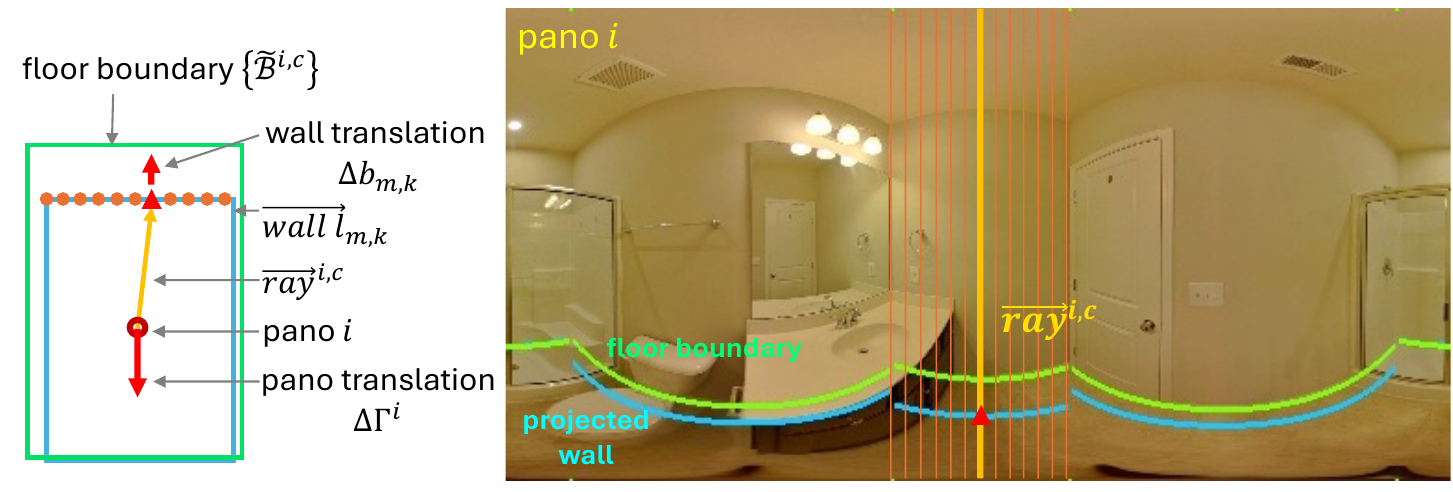}
    \caption{\footnotesize  Column-wise planar \textit{BA} module. Positional adjustment for walls and cameras are computed for each image column. At each column, the associated \textcolor{cyan}{wall} $l_{m,k}$ is projected with the current camera pose $E^i$. The adjustments are computed by comparing the projected point to the \textcolor{green}{floor boundary} $\Tilde{\mathcal{B}}^{i, c}$ value. The dense per-column adjustments are estimated in parallel with our \textit{BA} layer implementation.} 
    \label{fig:column_ba}
    \vspace{-24pt}
\end{center}
\end{figure}

Wide-baseline indoor reconstruction often suffers from a lack of robust matching features with sub-pixel accuracy. However, each wall can be observed by several image columns, with their floor boundary modeled as a line, as shown in Figure \ref{fig:column_ba}. \textit{BADGR} tackles the multi-view floor plan and pose reconstruction problem using a planar bundle adjustment (\textit{BA}), minimizing reprojection errors between the input predicted floor boundaries from image-based models and the projected wall positions. These errors are computed at the column level and are used to adjust the wall translation along its normal direction and the corresponding camera pose, optimized via a \textit{Levenberg-Marquardt} (\textit{LM}) step.

Noise in the floor boundaries can introduce errors in the final \textit{BA} results. 
To mitigate the noisy signal and incorporate learned layout-structural constraints, \textit{BADGR} integrates the planar \textit{BA} mechanism into a conditional denoising diffusion process. Instead of averaging column-wise wall and camera movements, the model encodes these dense signals and uses them to condition a transformer-based diffusion model. In this framework, posterior sampling is employed to generate the final adjustments, using raw \textit{BA} adjustments as inputs.
The transformer model predicts the \textit{xy} positions of camera poses and room vertices. To facilitate iterative interaction with planar \textit{BA}, an angle-constrained scene representation is used, along with a \textit{Layout Directional Regularizer} (\textit{LDR}) to form a bidirectional map between 2D room vertex positions and wall positions while maintaining fixed angles. The details of this process are explained in the following section.

\noindent\textbf{Architecture:}
Figure \ref{fig:Architecture} shows the architecture of our proposed optimization pipeline. Taking inspiration from \textit{HouseDiffusion} \cite{shabani2023housediffusion}, \textit{BADGR} has a diffusion model \cite{ho2020ddpm} based architecture with a truncated denoising inference process \cite{meng2021sdedit, zheng2022truncated} during training and inference. At each time $t$, our model takes a scene of 2D positions, i.e. \resizebox{0.27\hsize}{!}{$(\{\mathcal{T}^{i,(t)}\}, \{V_{m}^{(t)}\})$}, conditioned on: 1) the wall and panorama metadata encoded by an Entity Encoder discussed below, 
2) the floor boundary and adjustments embedding encoded by the Column Geometry Encoder, and generates an updated scene, i.e. \resizebox{0.32\hsize}{!}{$(\{\mathcal{T}^{i,(t-1)}\}, \{V_{m}^{(t-1)}\})$}, for time $t-1$.

The architecture consists of: (a) a planar \textit{BA} layer, (b) a Column Geometry Encoder module, where column-wise dense adjustments are encoded into a 1D embedding for each wall and panorama, (c) a Coordinate Encoder to embed 2D vectors for directional vector and \textit{xy} coordinates, d) an Entity Encoder to generate an entity embeddings for metadata of each wall and panorama, functioning similarly as a positional encoding, and (e) a self-attention Transformer denoiser to integrate the embeddings along with time step and to estimate new 2D positions.

\noindent\textbf{Entity Encoder:} Similar to the ``input conditions" from \textit{HouseDiffusion}, a size-D identity embedding is generated for each wall and camera using metadata, e.g. wall direction vector, one-hot room type, room id and vertex id for wall, and camera direction vector, one-hot camera ids for panorama. Wall and camera embeddings are generated with separate \textit{MLP} units. 

\vspace{-6pt}
% \begin{minipage}{0.8\linewidth} 

% \begin{figure}
\begin{savenotes}
\begin{algorithm}

\footnotesize
\caption{\footnotesize BA Optimization on Column $\mathcal{C}^{i, c}$}
\label{alg:ba}
\KwIn{Wall parameters $(b_{m,k}, \overrightarrow{v_{m,k}})$, camera 2D pose $\mathcal{T}^{i}$, $\mathcal{R}^{i}$, camera height $z^i$, and floor boundary $\Tilde{\mathcal{B}}^{i, c}$}
\KwOut{$\Delta b_{m,k}^{i, c}, \Delta \mathcal{T}_{m,k}^{i, c}$}

\resizebox{0.55\hsize}{!}{$\overrightarrow{ray}^{i, c}=(\mathcal{T}^{i}, [sin\mathcal{R}^{i}, cos\mathcal{R}^{i}])$}

\resizebox{0.45\hsize}{!}{$\overrightarrow{wall}_{m,k}=( \overrightarrow{v_{m,k}}, b_{m,k})$}

Global CS \footnote{Coordinate System (CS)}: \resizebox{0.6\hsize}{!}{${pt}^{i,c}_{m,k} \gets \textbf{Intersect}(\overrightarrow{ray}^{i, c}, \overrightarrow{wall}_{m,k})$}

Camera CS: \resizebox{0.7\hsize}{!}{${pt}^{i,c}_{m,k} \gets \textbf{GlobalToCam2D}({pt}^{i,c}_{m,k}, \mathcal{T}^{i}, \mathcal{R}^{i})$}

Projected boundary: $\hat{\mathcal{B}}^{i, c}_{m,k}=\textbf{Cam2DToPixel}({pt}^{i,c}_{m,k}, z^i)\textbf{.row} $

Reprojection error function: \resizebox{0.55\hsize}{!}{$\epsilon^{i,c}_{m,k}(b_{m,k}, \mathcal{T}^{i})\gets|\hat{\mathcal{B}}^{i, c}_{m,k}-\Tilde{\mathcal{B}}^{i, c}|$}

Jacobian matrix function: \resizebox{0.5\hsize}{!}{$\mathcal{J}^{i,c}_{m,k}(b_{m,k}, \mathcal{T}^{i})\gets\textbf{Jacrev}(\epsilon^{i,c}_{m,k})$}

Single-step \textit{LM}: \resizebox{0.8\hsize}{!}{$\Delta b_{m,k}^{i, c}, \Delta \mathcal{T}_{m,k}^{i, c} \gets \textbf{LM}(\epsilon^{i,c}_{m,k}, \mathcal{J}^{i,c}_{m,k},b_{m,k},\mathcal{T}^{i})$}

Final wall adjustment: \resizebox{0.55\hsize}{!}{$\Delta b_{m,k}^{i, c} \gets  \textbf{AdaptiveHuber}\text{\cite{barron2019adaptivehuber}}(\Delta b_{m,k}^{i, c}$})

Final pose adjustment: \resizebox{0.55\hsize}{!}{$\Delta \mathcal{T}_{m,k}^{i, c} \gets \textbf{AdaptiveHuber}(\Delta \mathcal{T}_{m,k}^{i, c}$})
\end{algorithm}
\end{savenotes}

% \end{figure}
% \end{minipage}
\vspace{-4pt}

\noindent\textbf{Planar \textit{BA} Layer:} \label{sec:ba} This layer generates camera and wall position adjustments by comparing the projected floor boundary from a given scene state, i.e. layouts and camera poses, with the input predicted floor boundary. The adjustments are computed densely on each image column with a \textit{LM} optimization algorithm, which is set to run a single step. The process is demonstrated in Figure \ref{fig:column_ba}. Compared to \textit{gradient descent} (\textit{GD}) based optimization, \textit{LM} is more efficient for convergence as it combines \textit{GD} with the \textit{Gauss-Newton} algorithm \cite{gavin2019levenberg}. \textit{LM} optimization requires the computation of the \textit{Jacobian} matrix from each adjusted parameter at each column position. 
To enable efficient training of the proposed learned \textit{BA} diffusion model, the \textit{LM}-adjusted parameters for each feature column are limited to three primary values: $\Delta b_{m,k}^{i, c}$ representing wall bias and $\Delta \mathcal{T}_{m,k}^{i, c}$ denoting the \textit{xy} displacement of the camera. This approach reduces computational complexity and memory overhead, enabling an efficient planar \textit{BA} implementation as a PyTorch layer. However, this constraint necessitates pre-assigning each image column to a designated global wall to streamline the reprojection process by directly referencing the corresponding wall and camera positions without occlusion checks for multiple walls. This initial column-to-wall assignment is established during scene initialization, using feature-matching models such as \textit{CovisPose} to ensure robust alignment prior to \textit{BADGR} optimization. The \textit{BA} layer is implemented with the differentiable non-linear optimization library \textit{Theseus} \cite{pineda2022theseus}; this library builds on PyTorch and applies a sparse solver with both CUDA and CPU implementations. See Algorithm \ref{alg:ba} for implementation details.

\noindent\textbf{Column Geometry Encoder:} This encoder is designed to compress the dense adjustments $(\Delta b_{m,k}^{i, c}, \Delta \mathcal{T}_{m,k}^{i, c})$ and the floor boundary $\{\Tilde{\mathcal{B}}^{i, c} \}$ into a 1D embedding for each wall and camera. The input array of shape $N_{p} \times N_{l} \times w \times 6$ is stored as three 2D vectors densely collected from feature columns of $N_{p}$ panoramas. Each column is assigned to at max a single wall. The resulting 1D embeddings, $N_{p} \times D$ for walls and $N_{l} \times D$ for cameras, are computed by compressing the three remaining axes. In our experiments, feature dimension $D=1024$.
Specifically, from a sparse array of shape $N_{p} \times N_{l} \times w \times 6$, the Column Geometry Encoder reduces the $w$ dimension by computing the projected mean and standard deviation vectors along their corresponding wall normal and wall vector axes, among the valid values along the $w$ axis. The Coordinate Encoder, which is based on \textit{learned Fourier Features} \cite{tancik2020fourier}, then encodes each stat vector into a size-D embedding $z_c$ by learning two sets of weights for mean and standard deviation vectors. The weights for mean vectors are also used to encode {$(\{\mathcal{T}^{i}\}, \{V_{m}\})$}. These features are concatenated and fed into a single \textit{Fully-Connected} (\textit{FC}) layer to generate a geometric guidance embedding of shape $N_{p} \times N_{l} \times 1024$. Since adjustments $(\Delta b_{m,k}^{i, c}, \Delta \mathcal{T}_{m,k}^{i, c})$ are relatively small numbers compared to $(b_{m,k}, \mathcal{T}^{i})$, We scale them by a factor of 100 prior to input to Column Geometry Encoder.

To further compress features at per-wall level, the entity embedding of cameras produced by the Entity Encoder is added into each of the $N_{l}$ array columns. 2-layer 1D convolutions \cite{van2016wavenet} followed by \textit{LeakyReLu} \cite{radford2015leakyrelu} and a single \textit{FC} layer are applied to produce a per-wall geometric guidance embedding of shape $N_{p} \times D$. Similarly, the entity embedding of walls is added into the $N_{p}$ array rows. A separate similar 1D convolutional network is used to generate per-camera geometric guidance embeddings of shape $N_{l} \times D$.

\noindent \textbf{Transformer Denoiser:}
 Four layers of self-attention mechanisms \cite{vaswani2017attention} are used to denoise the input scene and reason about the relationships between different scene entities and geometric guidance. The layers use input / output features of shape $(N_p + N_l)\times D$. The output features are fed into a \textit{FC} layer to produce \textit{xy} coordinates as final outputs for $(\{\mathcal{T}^{i}\}, \{V_{m}\})$. In each attention layer,  3 types of attention heads with different masking schemes, i.e. Component-wise Self Attention mask (CSA), Global Self Attention mask (GSA) and Relational Cross Attention mask (RSA), were applied, similar to \textit{HouseDiffusion} \cite{shabani2023housediffusion}. For the additional camera inputs, all cells related to valid cameras were left unmasked. Each masked type above contains 4 heads. The outputs of attention layers are summed and fed into add and norm layers. At the end, a single-layer \textit{MLP} is used to predict diffusion noise from current time stamps $t_n$ to $t_0$.

\noindent\textbf{Angle-Constrained Layouts:} These layouts are critical to connect the planar \textit{BA} module with the Transformer Denoiser in an iterative pipeline. The Planar \textit{BA} module processes layouts as inputs and outputs with line representation {$(\overrightarrow{v_{m,k}}$, $b_{m,k})$} of 1-\textit{DoF} walls, while the Transformer works with 2D coordinates $\{V_{m, k}\}$. Essentially, the Angle-Constrained Layouts enable two-way mappings between {$(\overrightarrow{v_{m,k}}$, $b_{m,k})$} and $\{V_{m, k}\}$, by defining half of the $\{V_{m, k}\}$ \textit{xy} coordinate values predicted from the Transformer Denoiser as irrelevant to the final layout prediction. With wall directions $\overrightarrow{v_{m,k}}$ as fixed vectors, the \textit{DoF} of $\{V_{m}\}$ are reduced by half using the $xy$ validity mask $\tau_{m,k}$ of the same shape as $\{V_{m}\}$. $\tau_{m,k}$ is generated in the scene initialization stage, along with the wall directional vector $\{\overrightarrow{v_{m,k}}\}$. The invalid values defined in $\tau_{m,k}$ will be overwritten by outputs of the proposed \textit{Layout Directional Regularizer} (\textit{LDR}).

The \textit{LDR} starts from a point $v_{m,k}$, with two valid $xy$ coordinates, and updates the invalid \textit{xy} positions of next vertex $v_{m,k+1}$, using wall direction $\overrightarrow{v_{m,k}}$, $v_{m,k+1}$ and validility $\tau_{m,k+1}$. The \textit{LDR} is applied around the loop of each layout polygon to update all vertices. Angle constrained walls allow $\textit{BADGR}$ to be trained to predict wall movement along the normal direction, while still using 2D coordinates to represent the layout for design simplicity. $\tau_{m,k}$ is also used as an input condition and a mask in L2 loss computation.

\noindent \textbf{Diffusion Model:}
\textit{BADGR} adopts the \textit{DDPM} process to learn to invert a diffusion process which adds noise to data with function $q(x)$ \cite{ho2020ddpm}. In the forward process, Gaussian noise is added to directly produce $x_t$ from $x_0$, same as \cite{ho2020ddpm}. We train the \textit{non-Markovian} diffusion processes, to predict noise from $x_t$ to $x_0$ and uses them to interpolate $x_{t-1}$ \cite{song2020_ddim}. This allows \textit{BADGR} to combine nonlinear optimization and diffusion without explicitly defining each step size. Loss for measuring view-consistency can be directly applied to the predicted $x_0$, which reuses the weighting scheme of a regular \textit{DDPM} model for different time stamp during training. For inference, \textit{probability flow ODE} \cite{song2020_score_sde} is used to iteratively denoise samples.

\noindent \textbf{Loss Function:} \textit{BADGR} is trained to perform \textit{BA} optimization, with 1) input conditions from dense column-wise \textit{BA} adjustments, and 2) a reprojection loss, similar to the traditional \textit{BA}, to regularize view-consistency. The loss function can be written as: 
\vspace{-2pt}
\begin{equation}
\resizebox{0.5\hsize}{!}{
$\mathcal{L}^{(t)} = \mathcal{L}_{L2}^{(t)} + \mathcal{W}_{proj} * \mathcal{L}_{proj}^{(t)}$
}
\end{equation}
% \vspace{-4pt}

\noindent where $\mathcal{L}_{L2}^{(t)}$ is the masked L2 reconstruction loss, and is only computed for valid \textit{xy} coordinates defined by $\tau_{m,k}$, from our Angle-constrained layouts. It's computed as: 
\begin{equation}
\resizebox{0.85\hsize}{!}{
$\mathcal{L}_{L2}^{(t)} = \lVert ((v_{m,k}^{pred}, \mathcal{T}^{i, pred}) - (v_{m,k}^{gt}, \mathcal{T}^{i, gt})*\tau_{m,k})\rVert_2$
}
\end{equation} 

\noindent $\mathcal{L}_{proj}^{(t)}$ is the layout-to-image re-projection loss, which is computed among all the columns with pre-assigned global wall for \textit{BA} adjustment using estimated scene at time 0.
\vspace{-3pt}
\begin{equation}
\begin{split}
\resizebox{0.35\hsize}{!}{
$\mathcal{L}_{proj}^{(t)} = \lVert \epsilon_{m,k}^{i,c,\widetilde{t=0}}\rVert_1$
}
\end{split}
\end{equation}
The process is described in steps 1-6 in Algorithm \ref{alg:ba} measured in pixel units. $\mathcal{W}^{proj}$ is the time independent weight for the projection loss, set    to $100$. 
\section{Experiments}
% Experiments are design using two types of floor plan data: 1) trained on a new FloorPlan-60K dataset and tested on ZInD \cite{cruz2021zillow}, which has similar data collection process and data distribution, 2) trained and tested on RPLAN dataset \cite{wu2019_rplan}.
We design experiments using two types of floor plan data. First, we employ an end-to-end pipeline involving coarse scene initialization followed by \textit{BADGR}, trained on a newly introduced FloorPlan-60K dataset and evaluated on ZInD \cite{cruz2021zillow}, which has similar data collection and distribution characteristics. Second, we assess \textit{BADGR} independently by training and testing it on the RPLAN dataset \cite{wu2019_rplan}, with controlled noise added to both layout and virtual camera poses. In the Supplementary, we report RPLAN-trained accuracy evaluated on ZInD to highlight \textit{BADGR}'s key ability to train on 2D schematic views and generalize across datasets. While it doesn’t perform as well as the FloorPlan60K-trained model, RPLAN-trained \textit{BADGR} still significantly outperforms our baseline approach.

% Although not as good as FloorPlan60K-trained model, RPLAN-trained \textit{BADGR} still beats our baseline approach significantly. 

% Significantly better than the SOTA, but not as good as trained on FP-60K.
% We also report accuracy trained with RPLAN and evaluated on ZInD in Supplementary, to demonstrate the \textit{BADGR}'s generalization ability.

\subsection{Experiments with FloorPlan-60K Data}
We use FloorPlan-60K, an extended version of ZInD \cite{cruz2021zillow}, generated through a similar production pipeline to provide the scale needed for our diffusion-based \textit{BA} training. FloorPlan-60K includes 68,147 floor plans, with an average of 8.7 rooms per plan and around 6.9 walls per room. Most walls align with Manhattan-world assumptions: 96.2\% are at 90 degrees, 3.0\% at 45 or 135 degrees, and 0.8\% form other angles. We have permission from Zillow to use this dataset, with a public release pending from the owner. At a minimum, model weights will be available for reproducibility. Our training stage uses only layouts and simulated poses, minimizing privacy concerns. All evaluations are conducted on the public ZInD test set, ensuring reproducibility. For end-to-end testing, we use panoramas from ZInD, starting with initial coarse room layouts and positions from the \textit{CoVisPose+} method (see section \ref{sec:baseline_models}), which serve as inputs to our main contribution: the learned-based global refinement, or \textit{BADGR}. Details on training and inference are provided in the Supplementary.

\begin{figure*}[t]
\begin{center}
    \includegraphics[width=0.83\linewidth]{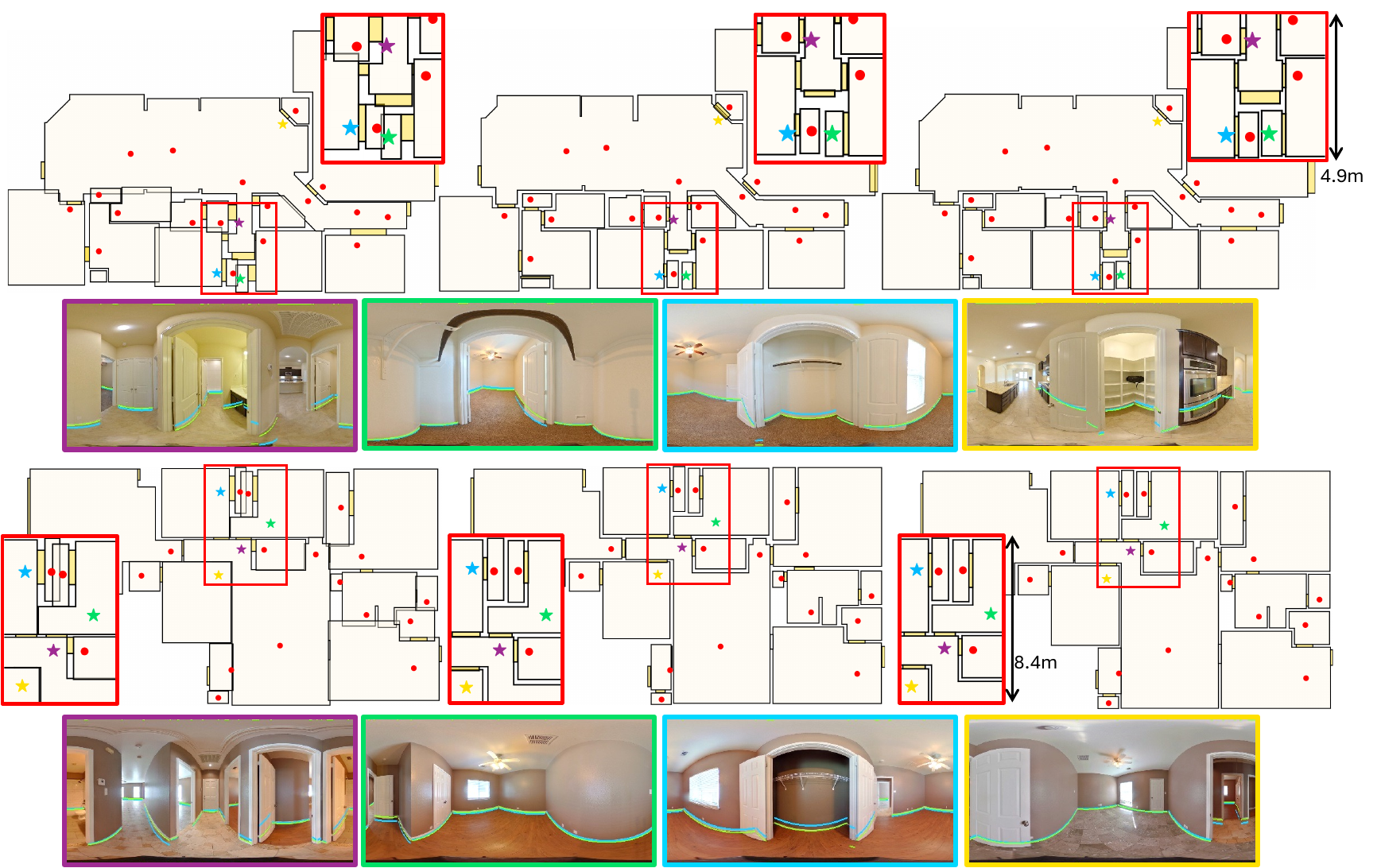}
    \vspace{-5pt}
    \caption{Qualitative results: top-down layouts and poses before (left), after \textit{BADGR} optimization (middle), and GT (right). The reprojected geometry, \textcolor{blue}{before} and \textcolor{green}{after} optimization, is shown in several images, highlighting the improved view-consistency, border colors indicate the capture positions. Example areas with significant improvements are \textcolor{red}{highlighted} and zoomed in. More examples in Supplementary.} 
    \label{fig:qualitative}
\end{center}
\vspace{-20pt}
\end{figure*}

\subsubsection{Baseline Models}
\label{sec:baseline_models}
% \textbf{\textit{CovisPose}} The modified \textit{CovisPose} model, see Section \ref{sec:init}, was run on every pair of straightened panoramas \cite{zhang2014panocontext} from a given floor. As a baseline method ``\textit{CovisPose}", \textit{Greedy Spanning Tree} \cite{nejatishahidin2023graph} was used to generate global poses. The modified \textit{CovisPose} model is applied to each pair of straightened panoramas \cite{zhang2014panocontext} for a given floor. \textbf{As a first baseline}, we start with the original \textit{CovisPose} method followed by a \textit{Greedy Spanning Tree} algorithm \cite{nejatishahidin2023graph} to generate global camera poses. \textbf{As a second baseline}, also the intial state of \textit{BADGR}, to further enhance pose accuracy, we introduce \textit{CovisPose+}, an improved version of \textit{CovisPose}, where vanishing point snapping is incorporated into the pairwise pose estimation step before applying the \textit{Greedy Spanning Tree}.

\textbf{\textit{CovisPose}:} The modified \textit{CovisPose} model (see Section \ref{sec:init}) was applied to each pair of straightened panoramas \cite{zhang2014panocontext} from a given floor. For the first baseline, we used the original \textit{CovisPose} method followed by the \textit{Greedy Spanning Tree} (\textit{GST}) algorithm \cite{nejatishahidin2023graph} to generate global poses. For the second baseline, which is also the initial state of \textit{BADGR}, we introduced \textit{CovisPose+}, an improved version of \textit{CovisPose} that incorporates vanishing point snapping in the pairwise pose estimation step before applying the \textit{GST}.

\noindent \textbf{\textit{BA-Only}:} This method applies only the planar \textit{BA} layer to refine camera and wall positions iteratively, starting from the coarse initialization in Section \ref{sec:init}. First, per-column adjustments are made as described in steps 8-10 of Algorithm \ref{alg:ba}. These adjustments are then grouped by wall and camera, and averaged to produce the final position updates. To resolve conflicts during averaging, majority voting selects the dominant direction for each adjustment. The \textit{LM} optimization is run for 100 iterations with a step size 2.5 times larger than the original.
% This approach only applies \textit{Bundle Adjustment Layer} on the input scene to update both camera positions and wall positions \cite{agarwal2010bundle, pineda2022theseus}. Per-column dense adjustments are computed first with step 8-10 in Algorithm \ref{alg:ba}. Then the dense adjustments are grouped by their wall and camera identities and averaged as the final per-wall and per-camera positional adjustments. To reduce conflicting gradients when averaging dense adjustment, majority voting is used to select only the positive or negative value group. The \textit{LM} optimization is run for 100 steps with 2.5X of the original step size. The output, per-column adjustments, are averaged according to their camera and wall identities.
% This method applies only the planar \textit{BA} layer to iteratively refine both camera and wall positions from the input coarse initialization, described in Section \ref{sec:init}. Initially, per-column adjustments are computed following steps 8-10 of Algorithm \ref{alg:ba}. These dense adjustments are then grouped by wall and camera identities, with an averaging step that consolidates them into the final positional updates for each wall and camera. To mitigate conflicting gradients during averaging, majority voting is employed to select the dominant adjustment direction, either positive or negative. The \textit{LM} optimization is executed over 100 iterations with an expanded step size of 2.5 times the original. The resulting per-column adjustments are averaged by camera and wall identity for coherent scene alignment.

\noindent \textbf{Other Methods:} \textit{CovisPose} \cite{hutchcroft2022covispose} outperforms point-matching \textit{SfM}-based methods \cite{lowe2004distinctive, sun2021loftr} and recent learning-based approaches like \textit{DirectionNet} \cite{chen2021_directionnet} and \textit{SALVe} \cite{lambert2022salve} in wide-baseline indoor pose estimation. While \textit{GraphCovis} estimates poses from three to five panoramas, it cannot be applied directly to our test cases, which typically involve more than 10 panoramas (see Table \ref{tab:combined_zind}). Additionally, the lower pose errors shown in \cite{nejatishahidin2023graph} are limited to configurations with only three panoramas, unsuitable as a comprehensive baseline. Similarly, \textit{PanoPose} \cite{tu2024panopose} estimates relative poses between pairs of panoramas but lacks publicly available code for reproducibility.

\vspace{-4pt}
\subsubsection{Results}
We evaluate models on panorama subsets with varying input densities, as shown in Table \ref{tab:combined_zind}. Pose errors (mean, median, standard deviation, and p90) are computed after aligning predicted and ground truth pose graphs using RANSAC \cite{nejatishahidin2023graph}, with similar metrics for layout errors. The number of images per room used in experiments is defined with partial rooms, where a complete ZInD room contains one or more partial rooms \cite{cruz2021zillow}.

% Models are evaluated using subsets of panoramas with different input densities, as shown in Table \ref{tab:combined_zind}. The number of images per room used in experiments is defined with partial rooms, with a complete ZInD room containing one or more partial rooms. The pose error metrics include the mean (Mn), median (Med), standard deviation (Std), and 90th percentile (p90) of absolute translation errors and absolute rotation errors after aligning the predicted and ground truth pose graphs using \textit{RANSAC} \cite{nejatishahidin2023graph}. Similar metrics for layout errors are reported after aligning the predicted floor plan vertices with the ground truth using \textit{RANSAC} \cite{nejatishahidin2023graph}.

% \usepackage{multirow}
\begin{table}[t]
\centering
% \vspace{5mm}
\caption{Pose and layout error from predictions on ZInD dataset. Note that \textit{BA}-Only and \textit{BADGR} don't optimize camera rotations, hence share the same rotation errors as \textit{CovisPose+}.}
\label{tab:combined_zind}
\vspace{-5pt}
\scalebox{0.55}{
\begin{tabular}{cc|cccccc|cccc|c}
\hline
\multirow{2}{*}{\begin{tabular}[c]{@{}c@{}}Imgs/\\ Rm\end{tabular}} & \multirow{2}{*}{Methods} & \multicolumn{3}{c}{Pose Rot.($^\circ$)}       & \multicolumn{3}{c|}{Pose Transl.(cm)}                                           & \multicolumn{4}{c|}{Visible walls (cm)}                    & \multirow{2}{*}{\begin{tabular}[c]{@{}c@{}}\#walls \\ \#panos\end{tabular}} \\ \cline{3-12}
                                                                    &                          & Mn            & Med           & Std           & Mn                       & Med                      & Std                       & Mn           & Med          & Std          & p90           &                                                                             \\ \hline
\multirow{4}{*}{0.6}                                                & \textit{CovisPose}       & 1.83          & 1.25          & 1.48          & 22.9                     & 17.4                     & 13.4                      & 14.0         & 8.0          & 12.8         & 32.4          & \multirow{4}{*}{\begin{tabular}[c]{@{}c@{}}59.4\\ 8.0\end{tabular}}         \\
                                                                    & \textit{CovisPose+}      & \textbf{0.24} & \textbf{0.20} & \textbf{0.30} & 20.7                     & 15.7                     & 12.0                      & 11.5         & 6.9          & 10.5         & 26.0          &                                                                             \\
                                                                    & \textit{BA-Only}         & \textbf{0.24} & \textbf{0.20} & \textbf{0.30} & 19.1                     & 12.2                     & 10.9                      & 12.8         & 6.8          & 11.9         & 29.3          &                                                                             \\
                                                                    & \textit{BADGR}           & \textbf{0.24} & \textbf{0.20} & \textbf{0.30} & \textbf{12.2}            & \textbf{9.5}             & \textbf{7.2}              & \textbf{7.1} & \textbf{4.5} & \textbf{6.7} & \textbf{15.3} &                                                                             \\ \hline
\multirow{4}{*}{1}                                                  & \textit{CovisPose}       & 1.88          & 1.36          & 1.40          & 23.0                     & 18.0                     & 13.2                      & 15.0         & 9.0          & 13.0         & 39.1          & \multirow{4}{*}{\begin{tabular}[c]{@{}c@{}}65.7\\ 10.4\end{tabular}}        \\
                                                                    & \textit{CovisPose+}      & \textbf{0.26} & \textbf{0.21} & \textbf{0.30} & 19.7                     & 15.6                     & 11.8                      & 12.3         & 7.0          & 10.8         & 27.4          &                                                                             \\
                                                                    & \textit{BA-Only}         & \textbf{0.26} & \textbf{0.21} & \textbf{0.30} & \multicolumn{1}{l}{17.6} & \multicolumn{1}{l}{12.6} & \multicolumn{1}{l|}{12.0} & 12.6         & 6.7          & 11.4         & 27.5          &                                                                             \\
                                                                    & \textit{BADGR}           & \textbf{0.26} & \textbf{0.21} & \textbf{0.30} & \textbf{11.2}            & \textbf{8.8}             & \textbf{6.5}              & \textbf{7.0} & \textbf{4.6} & \textbf{6.6} & \textbf{15.8} &                                                                             \\ \hline
\multirow{4}{*}{2}                                                  & \textit{CovisPose}       & 1.90          & 1.49          & 1.31          & 22.3                     & 17.7                     & 12.6                      & 14.7         & 8.8          & 12.5         & 33.6          & \multirow{4}{*}{\begin{tabular}[c]{@{}c@{}}65.9\\ 16.0\end{tabular}}        \\
                                                                    & \textit{CovisPose+}      & \textbf{0.26} & \textbf{0.19} & \textbf{0.30} & 17.9                     & 14.5                     & 10.2                      & 12.0         & 7.0          & 9.0          & 26.0          &                                                                             \\
                                                                    & \textit{BA-Only}         & \textbf{0.26} & \textbf{0.19} & \textbf{0.30} & 14.3                     & 10.8                     & 9.2                       & 10.5         & 6.2          & 9.9          & 23.3          &                                                                             \\
                                                                    & \textit{BADGR}           & \textbf{0.26} & \textbf{0.19} & \textbf{0.30} & \textbf{10.7}            & \textbf{8.9}             & \textbf{6.0}              & \textbf{6.4} & \textbf{4.4} & \textbf{5.9} & \textbf{13.9} &                                                                             \\ \hline
\multirow{4}{*}{3}                                                  & \textit{CovisPose}       & 1.70          & 1.19          & 1.23          & 21.5                     & 17.1                     & 11.9                      & 14.3         & 8.5          & 12.1         & 31.8          & \multirow{4}{*}{\begin{tabular}[c]{@{}c@{}}66.5\\ 18.3\end{tabular}}        \\
                                                                    & \textit{CovisPose+}      & \textbf{0.23} & \textbf{0.19} & \textbf{0.27} & 18.1                     & 14.8                     & 10.5                      & 12.3         & 7.2          & 10.2         & 26.9          &                                                                             \\
                                                                    & \textit{BA-Only}         & \textbf{0.23} & \textbf{0.19} & \textbf{0.27} & 13.4                     & 10.7                     & 8.2                       & 10.6         & 6.2          & 9.0          & 22.1          &                                                                             \\
                                                                    & \textit{BADGR}           & \textbf{0.23} & \textbf{0.19} & \textbf{0.27} & \textbf{10.6}            & \textbf{8.9}             & \textbf{6.0}              & \textbf{6.6} & \textbf{4.3} & \textbf{6.0} & \textbf{14.6} &                                                                             \\ \hline
\end{tabular}
}
\vspace{-14pt}
\end{table}

% Models are evaluated using subsets of panoramas sampled with different input density settings from each floor, reported in Table \ref{tab:combined_zind}. The number of panoramas per room is defined with partial rooms, where a complete ZInD room contains one or multiple partial rooms. Metrics measuring pose errors reported include: mean (Mn), median (Med), standard deviation (Std), 90th percentile (p90) of absolute translation errors (meters) and absolute rotation errors (degrees) after aligning the predicted pose graph with the ground truth pose graph using \textit{RANSAC} \cite{nejatishahidin2023graph}. For layout errors, similar metrics are reported after aligning the predicted floor plan vertices with the ground truth using \textit{RANSAC} \cite{nejatishahidin2023graph}. 

As shown in Table \ref{tab:combined_zind}, the \textit{BA-Only} baseline reduces pose and layout errors from \textit{CovisPose+} when input density is sufficient, e.g. more than one image per room, with accuracy improving as input images increase. \textit{BADGR} follows a similar trend, consistently achieving lower layout and pose errors across different input densities, including in extreme-baseline scenario with 0.6 images per room. This highlights the effectiveness of \textit{BADGR}'s learned layout-structural constraints and its understanding of global context when compared to the view-consistency-only approach from the \textit{BA-Only} baseline. 

\subsection{Experiments with RPLAN}

The RPLAN dataset \cite{wu2019_rplan} is used to evaluate \textit{BADGR} in controlled noise experiments during training and testing. RPLAN is a large-scale dataset of real residential floor plans, each with 4 to 8 Manhattan rooms, spanning 65 to 120 square meters, but without real-world scale. After preprocessing \cite{shabani2023housediffusion} and aligning door edges with walls, we use 57,303 plans for training and 19,000 for testing. The \textit{BADGR} model is trained for 20 diffusion steps with an embedding size $D=512$, handling up to 100 walls and 15 cameras, using the training procedure from FloorPlan-60K. Since RPLAN lacks real indoor images, we use simulated camera poses and rendered floor boundaries for evaluation. Noisy layouts and poses are created by adding Gaussian noise (mean = 0, standard deviation = 3.3\%) to the ground truth wall and camera positions. These inputs are normalized to a range of [-1, 1], and error distances (shown in Table \ref{tab:rplan}) are reported in this normalized space.

\begin{table}[t]
\centering
\vspace{0pt}
% \caption{\footnotesize Pose and layout errors computed from RPLAN dataset with simulated panorama poses and rendered floor boundaries. Since RPLAN floors are all under 120 square meters, the reported distance error in percentage can be mapped to real metric scale by roughly 1\% to 0.1 meter.}
\caption{Pose and layout errors from predictions on RPLAN dataset. Since RPLAN floors are under 120 square meters, the reported distance error in percentage roughly translates to 1\% to 0.1 meter in real scale.}
\vspace{-6pt}
\scalebox{0.64}{
\begin{tabular}{ccccccccccc}
\hline
\multirow{2}{*}{\begin{tabular}[c]{@{}c@{}}Imgs /\\ Rms\end{tabular}} & \multirow{2}{*}{\begin{tabular}[c]{@{}c@{}}State / \\ Method\end{tabular}} & \multicolumn{4}{c}{Pose Err Dist (\%)}                        & \multicolumn{4}{c}{Visible Layout Err Dist (\%)}              & \multirow{2}{*}{\begin{tabular}[c]{@{}c@{}}\# walls\\ \# panos\end{tabular}} \\ \cline{3-10}
                                                                      &                                                                            & Mn            & Med           & Std           & p90           & Mn            & Med           & Std           & p90           &                                                                              \\ \hline
\multirow{3}{*}{1}                                                    & Start                                                                      & 3.15          & 2.94          & 1.56          & 4.98          & 1.69          & 1.46          & 1.25          & 3.42          & \multirow{3}{*}{\begin{tabular}[c]{@{}c@{}}46.7\\ 8.0\end{tabular}}          \\
                                                                      & \textit{BA-Only}                                                           & 0.58          & 0.41          & 0.31          & 0.93          & 0.62          & 0.30          & 0.73          & 1.25          &                                                                              \\
                                                                      & \textit{\textbf{BADGR}}                                                    & \textbf{0.35} & \textbf{0.29} & \textbf{0.18} & \textbf{0.55} & \textbf{0.33} & \textbf{0.15} & \textbf{0.38} & \textbf{0.58} &                                                                              \\ \hline
\multirow{3}{*}{2}                                                    & Start                                                                      & 3.21          & 2.93          & 1.59          & 5.20          & 1.70          & 1.45          & 1.26          & 3.44          & \multirow{3}{*}{\begin{tabular}[c]{@{}c@{}}46.7\\ 15.0\end{tabular}}         \\
                                                                      & \textit{BA-Only}                                                           & 0.62          & 0.31          & 0.38          & 1.09          & 0.58          & 0.23          & 0.73          & 1.11          &                                                                              \\
                                                                      & \textit{\textbf{BADGR}}                                                    & \textbf{0.34} & \textbf{0.23} & \textbf{0.17} & \textbf{0.54} & \textbf{0.27} & \textbf{0.13} & \textbf{0.33} & \textbf{0.47} &                                                                              \\ \hline
\end{tabular}
}
\label{tab:rplan}
\vspace{-0pt}
\end{table}

\begin{table}[t]
\centering
\vspace{-6pt}
\caption{Errors from predictions on RPLAN dataset with simulated noise in floor boundaries. First column contains input density, the chance and max scale of noise added to each visible wall.}
\vspace{-4pt}
\scalebox{0.66}{
\begin{tabular}{cccccccccc}
\hline
\multirow{2}{*}{\begin{tabular}[c]{@{}c@{}}Imgs/Rms\\ \%noise\end{tabular}}          & \multirow{2}{*}{\begin{tabular}[c]{@{}c@{}}State/ \\ Method\end{tabular}} & \multicolumn{4}{c}{Pose Err Dist (\%)}                        & \multicolumn{4}{c}{Visible Layout Err Dist (\%)}              \\ \cline{3-10} 
                                                                                     &                                                                           & Mn            & Med           & Std           & p90           & Mn            & Med           & Std           & p90           \\ \hline
\multirow{3}{*}{\begin{tabular}[c]{@{}c@{}}1\\ 5\% chance\\ 2\%scale\end{tabular}}   & Start                                                                     & 3.17          & 2.87          & 1.54          & 4.98          & 1.69          & 1.46          & 1.24          & 3.40          \\
                                                                                     & \textit{BA-Only}                                                          & 1.47          & 0.60          & 1.19          & 2.70          & 2.22          & 0.52          & 2.81          & 2.53          \\
                                                                                     & \textit{\textbf{BADGR}}                                                   & \textbf{0.59} & \textbf{0.42} & \textbf{0.35} & \textbf{0.98} & \textbf{0.49} & \textbf{0.27} & \textbf{0.50} & \textbf{0.95} \\ \hline
\multirow{3}{*}{\begin{tabular}[c]{@{}c@{}}1\\ 10\% chance\\ 2\%scale\end{tabular}}  & Start                                                                     & 3.17          & 2.92          & 1.57          & 4.99          & 1.69          & 1.45          & 1.24          & 3.41          \\
                                                                                     & \textit{BA-Only}                                                          & 1.63          & 0.79          & 1.25          & 2.93          & 1.95          & 0.62          & 3.07          & 2.67          \\
                                                                                     & \textit{\textbf{BADGR}}                                                   & \textbf{0.64} & \textbf{0.46} & \textbf{0.36} & \textbf{1.04} & \textbf{0.51} & \textbf{0.29} & \textbf{0.52} & \textbf{1.00} \\ \hline
\multirow{3}{*}{\begin{tabular}[c]{@{}c@{}}2\\ 5\% chance\\ 2\% scale\end{tabular}}  & Start                                                                     & 3.20          & 2.86          & 1.58          & 5.18          & 1.69          & 1.45          & 1.24          & 3.40          \\
                                                                                     & \textit{BA-Only}                                                          & 1.48          & 0.58          & 1.19          & 2.49          & 1.24          & 0.45          & 2.69          & 2.23          \\
                                                                                     & \textit{\textbf{BADGR}}                                                   & \textbf{0.65} & \textbf{0.41} & \textbf{0.33} & \textbf{1.05} & \textbf{0.43} & \textbf{0.22} & \textbf{0.45} & \textbf{0.80} \\ \hline
\multirow{3}{*}{\begin{tabular}[c]{@{}c@{}}2\\ 10\% chance\\ 2\% scale\end{tabular}} & Start                                                                     & 3.16          & 2.84          & 1.57          & 5.11          & 1.69          & 1.45          & 1.26          & 3.41          \\
                                                                                     & \textit{BA-Only}                                                          & 1.57          & 0.74          & 1.22          & 2.74          & 2.16          & 0.55          & 0.60          & 2.51          \\
                                                                                     & \textit{\textbf{BADGR}}                                                   & \textbf{0.72} & \textbf{0.46} & \textbf{0.38} & \textbf{1.17} & \textbf{0.56} & \textbf{0.25} & \textbf{0.57} & \textbf{0.87} \\ \hline
\end{tabular}
}
\vspace{-12pt}
\label{tab:rplan_noised}
\end{table}

\noindent \textbf{Simulated Noise in Floor Boundary Inputs} We added noise to each rendered floor boundary to simulate bias from boundary prediction models, caused by factors like occlusion and limited training data. Before rendering the floor boundary for each simulated camera position, we randomly translate each visible wall along its normal direction, adjusting the opposite side to avoid self-intersection. The walls altered are selected by chance. Noise follows a uniform distribution with a max scale, and is applied independently for each rendering. The resulting floor boundaries guide the \textit{BADGR} denoising process during testing.

% We added noise to each rendered floor boundary input to simulate the systematic bias from boundary prediction models due to factors like occlusion and limited training examples. Specifically, before rendering the floor boundary from each simulated panorama positions, each visible wall can be shifted by a random distance along its normal direction, with the opposite side also adjusted to prevent self-intersection. The noise generation is defined by a probability and a uniform distribution with a max noise scale. This noise is generated independently for each rendering. The resulting floor boundaries guide the \textit{BADGR} denoising process in test cases.

With a maximum noise level of 2\% (about 20 cm), both \textit{BADGR} and \textit{BA-Only} show increased pose and layout errors compared to no noise. However, \textit{BADGR} is significantly less impacted by input perturbations than \textit{BA-Only}. In terms of absolute distance error for poses and layouts, \textit{BADGR} achieves lower errors compared to the ZInD test case when approximate scale is applied, likely due to the simpler structure of RPLAN floor plans.

% We added noise to each rendered floor boundary input to simulate positional bias errors near walls due to factors like occlusion and limited training examples. Specifically, before rendering the floor boundary from each simulated panorama position, each visible wall can be shifted by a random distance along its normal direction, where the opposite side of wall is also modified to avoid self-intersection. This noise is generated independently for each rendering. The rendered floor boundaries then guide the \textit{BADGR} denoising process in test cases.

% We also added noise into the each rendered floor boundary inputs, to simulate the errors when the estimated floor boundary contains a positional bias for a wall, due to reasons like occlusion and a lack of similar training examples. Specifically, before rendering the floor boundary from each simulated panorama position, each visible wall has a chance to be shifted by a randomly generated distance along the normal direction. The noise simulation is generated independently for different panorama positions. The rendered floor boundaries are used to condition the \textit{BADGR} denoising process for test cases.
% \input{tables/rplan_noised_reproj}

% \noindent\textbf{Cross-dataset Evals} likely in supplementary. Training slow

% \subsubsection{Analysis with Noisy Inputs}
% \noindent\textbf{What if input number of walls are off?}

% \noindent\textbf{What if input depths are noisy? Biased?}

\subsection{Ablation Studies}
We demonstrate the impact of \textit{BADGR} in recovering poses and layouts by comparing to four models and training three \textit{BADGR} variants in Table \ref{tab:ablation}. Our conditional diffusion model \textit{BADGR} outperforms guided diffusion model \textit{BA+DM} in both tasks. \textit{BADGR} trained with \textit{BA} inputs only has a larger improvement compared with \textit{BADGR} trained with reprojection loss only. We also found that although having higher errors from the first four models, the resulting layouts are mostly plausible looking, with improved layout and pose accuracy from the starting point.  
% We also observed that without reprojection loss, \textit{BADGR} requires 30\% more iterations to converge with the similar loss curve.

% \textbf{BADGR vs guided diffusion vs BA only}
\begin{table}[t]
\centering
\vspace{0pt}
\caption{Ablation analysis on different variants of \textit{BADGR}. All models are trained on FloorPlan-60k datasets and tested on ZInD test set. Diffusion model (\textit{DM}) is \textit{BADGR} trained without planar \textit{BA} layer and reprojection loss.  \textit{BA}+\textit{DM} is a guided diffusion model, where \textit{BA} adjustment is added to the diffusion adjustment from \textit{DM} above without \textit{BA} conditioning.}
\vspace{-6pt}
\scalebox{0.6}{
\begin{tabular}{cc|c|cccc|cccc}
\hline
\multicolumn{2}{c|}{}                                                                                                                           & Imgs/Rms                & \multicolumn{4}{c|}{1}                                         & \multicolumn{4}{c}{2}                                          \\ \cline{3-11} 
\multirow{2}{*}{\begin{tabular}[c]{@{}c@{}}BA \\ Inputs\end{tabular}} & \multirow{2}{*}{\begin{tabular}[c]{@{}c@{}}Reproj \\ Loss\end{tabular}} & \multirow{2}{*}{Method} & \multicolumn{2}{c}{Vis. walls (cm)} & \multicolumn{2}{c|}{Pose (cm)}    & \multicolumn{2}{c}{Vis. walls (cm)} & \multicolumn{2}{c}{Pose (cm)}     \\
                                                                      &                                                                         &                         & Mn             & Std            & Mn            & Std          & Mn             & Std            & Mn            & Std          \\ \hline
\xmark                                                                & \xmark                                                                  & \textit{DM}         & 8.6            & 8.3            & 26.7          & 17.4         & 8.6            & 8.1            & 25.7          & 17.7         \\
\xmark                                                                & \cmark                                                                  & \textit{BADGR}          & 8.4            & 8.4            & 23.6          & 14.1         & 8.3            & 8.1            & 24.4          & 14.3         \\
\xmark                                                                & \xmark                                                                  & \textit{BA} + \textit{DM}    & 8.6            & 9.0            & 15.0          & 10.0         & 8.0            & 8.3            & 13.4          & 8.7          \\
\cmark                                                                & \xmark                                                                  & \textit{BADGR}          & 7.2            & 6.6            & 12.0          & 6.8          & 6.9            & 6.4            & 11.3          & 6.5          \\
\cmark                                                                & \cmark                                                                  & \textit{BADGR}                   & \textbf{7.0}   & \textbf{6.6}   & \textbf{11.2} & \textbf{6.5} & \textbf{6.4}   & \textbf{5.9}   & \textbf{10.7} & \textbf{6.0} \\ \hline
\end{tabular}
}
\label{tab:ablation}
\vspace{-14pt}
\end{table}

\subsection{Qualitative Results}
    As shown in Figure \ref{fig:qualitative}, BADGR improves layouts and view-consistency, even in extreme cases with minimal visual overlap, from a coarsely initialized scene. \textit{BADGR} is able to learn the physical constraints from training data and correct issues like overlapping room layouts and varying wall thicknesses. While pose errors are not easily visualized in a top-down view, the image view reveals these inaccuracies and highlights the substantial improvements in both poses and layouts achieved by \textit{BADGR}. Further evaluation of reprojection errors is provided in the Supplementary.
    % As shown in Figure \ref{fig:qualitative}, \textit{BADGR} improves layouts and enforces view-consistency from a coarsely initialized scene, even in challenging extreme-baseline reconstructions \cite{shabani2021extreme} with featureless scenes and minimal visual overlap among input images. \textit{BADGR} is able to learn the physical constraints from training data and correct issues like overlapping room layouts and varying wall thicknesses. While pose errors are not easily visualized in a top-down view, the image view reveals these inaccuracies and highlights the substantial improvements in both poses and layouts achieved by \textit{BADGR}. Further evaluation of reprojection errors is provided in the Supplementary.

\vspace{-2pt}

% As shown in Figure \ref{fig:qualitative}, \textit{BADGR} improves layouts and enforces view-consistency from a coarsely initialized scene. Both examples can be described as extreme-baseline reconstruction \cite{shabani2021extreme} with featureless scenes and little visual overlaps among input images. \textit{BADGR} improved input conditions which aren't aligned with training examples, such as overlapping room layouts and oversized wall thickness. Pose errors are difficult to visualize from topdown view. The image view demonstrates these errors and the drastic improvements on both poses and layouts from \textit{BADGR}. More evaluation on reprojection errors can be found in Supplementary.

% \noindent \textbf{Different training techniques: with / without drop out.}

\section{Conclusion}
% We introduce a novel diffusion model \textit{BADGR}, that integrates reconstruction and \textit{BA} optimization to refine pose and layout from floor boundaries derived from dozens of images. This is the first learning-based model to achieve wide-baseline, full-scale, view-consistent floor plan reconstruction from up to 30 images. We demonstrate improved reconstruction accuracy by optimizing view consistency with planar BA and by incorporating learned layout-structural constraints through diffusion modeling. \textit{BADGR} is trained exclusively on floor plan data only, allowing it to handle large scenes with many cameras, benefit from abundant training data, and support flexible data augmentation. Additionally, \textit{BADGR} demonstrates the advantages of combining a nonlinear optimization process with a denoising diffusion process with a straightforward training process, which learns constraints, and models observation errors and relations from training data.

We present \textit{BADGR}, a novel diffusion model that unites layout reconstruction with \textit{BA}-style optimization, refining coarse poses and layouts, such as those derived from multiple 360° panoramas using methods like \cite{hutchcroft2022covispose,lambert2022salve,tu2024panopose}, into accurate, view-consistent floor plans. This is the first learning-based approach designed to handle full-scale indoor environments with up to 30 capture points, achieving enhanced spatial coherence through an integrated approach combining planar \textit{BA} with diffusion-based structural constraints. Trained exclusively on schematic floor plans, \textit{BADGR} adeptly addresses complex layouts and supports robust data augmentation techniques, including simulated camera poses. By leveraging a conditional diffusion model to guide nonlinear optimization, \textit{BADGR} learns structural constraints and models spatial relationships and observation error, all through an efficient and effective training process.

\vspace{6pt}
\noindent
{\bf Acknowledgments.}
We are grateful to Sing Bing Kang for discussions and paper edits on this work. In addition, we appreciate Zillow Group on providing the FloorPlan-60K dataset throughout this research.     

{
    \small
    \bibliographystyle{ieeenat_fullname}
    \bibliography{main}
}

\clearpage
\onecolumn
\maketitlesupplementary
\setcounter{section}{0}
% \clearpage
\setcounter{page}{1}

\section{Implementation Details with FloorPlan-60K Data}
\label{sec:zillow_train}

\textit{BADGR} is trained using a 2D `cleanup' layer of floor plans, where larger spaces are represented by unions of multiple partially annotated room shapes, following the annotation approach of ZInD \cite{cruz2021zillow}. Panorama poses are randomly sampled within each room. For each input image, \textit{BADGR} simulates data with a CUDA-based 1D renderer, given floor plan layouts and a sampled camera pose. The renderer operates on connected rooms through doors, omitting door polygons and matching wall segments along the front and back planes. Random masking is applied on $\{\hat{\mathcal{B}^i}\}$ and $\mathcal{M}$ to occasionally bypass the BA layer for selected image columns. During diffusion training, scenes are rotated by $[0^{\circ}, 90^{\circ}, 180^{\circ}, 270^{\circ}]$. For evaluation, we use the ZInD test set (275 floor plans), with initial scenes, floor boundary depths, and column-to-wall assignments estimated from real panorama images, as detailed in Section 8.1 of the main paper.

\textit{BADGR} has a capacity of 300 walls and 30 panoramas, which is selected to accommodate 99\% of the floor plans from FloorPlan-60K data. It is trained with a batch size of 48, and with a learning rate of $10^{-4}$ for 140 epochs, $10^{-5}$ for 50 epochs, and $10^{-6}$ for 50 epochs by stepwise decay. \textit{BADGR} is trained for the last 20 steps of a 1000-step diffusion process, using a second-moment schedule sampler for time $t$. \textit{Ordinary Differential Equations} (\textit{ODE}) sampling \cite{song2020_score_sde} is used during the \textit{BADGR} inference process. Training peaks at 55GB GPU memory usage on a single GPU. During inference, \textit{BADGR} processes a batch size of 1 in approximately 25 seconds on a CPU-only Apple M1 MacBook Pro with 32GB of memory, and around 4.0 secs on an A100 GPU.

\section{Cross dataset training and validation}
 We additionally trained a \textit{BADGR} model of a max capacity of 300 walls and 30 cameras with the RPLAN training set, and with a similar settings of sampling camera positions for generating simulated floor boundaries and column-to-wall assignments. This model is evaluated on the ZInD test set. The results are listed in Table \ref{tab:cross_dataset} alongside with existing results for comparison. Although \textit{BADGR} trained with RPLAN dataset doesn't produce similar or higher accuracy than \textit{BADGR} trained with FloorPlan-60K dataset, it still outperform the \textit{CovisPose+} and \textit{BA-Only} baselines. This trend is expected as RPLAN contains Manhattan floors only and overall have less rooms and panoramas during training.

% \usepackage{multirow}
\begin{table}[H]
\vspace{-6pt}
% \caption{ Pose and layout error tested on ZInD dataset, trained with different datasets. Row 3 of each block is in addition to the main paper. Mn, Med, Std stands for mean, median and standard deviation. We also report the 90th percentile (p90) of the absolution translation errors from the estimated camera poses.}
\caption{Pose and layout errors tested on the ZInD dataset, trained with various datasets. Row 3 of each block presents additional results compared to the main paper. Mn, Med, and Std denote mean, median, and standard deviation, respectively. We also report the 90th percentile (p90) of the absolute translation errors for the estimated camera poses.}
\label{tab:cross_dataset}
\centering
\scalebox{0.8}{
\begin{tabular}{ccc|cccc|cccc}
\hline
\multirow{2}{*}{\begin{tabular}[c]{@{}c@{}}Imgs/\\ Rm\end{tabular}} & \multirow{2}{*}{Methods} & \multirow{2}{*}{Training Set} & \multicolumn{4}{c|}{Camera Translation(cm)}                 & \multicolumn{4}{c}{Visible walls (cm)}                     \\
                                                                    &                          &                               & Mn            & Med          & Std          & p90           & Mn           & Med          & Std          & p90           \\ \hline
\multirow{4}{*}{0.6}                                                & \textit{CovisPose+}      & ZInD                          & 20.7          & 15.7         & 12.0         & 33.6          & 11.5         & 6.9          & 10.5         & 26.0          \\
                                                                    & \textit{BA Only}         & N/A                           & 19.1          & 12.2         & 10.9         & 32.7          & 12.8         & 6.8          & 11.9         & 29.3          \\
                                                                    & \textit{BADGR}           & RPLAN                         & 14.7          & 11.4         & 8.0          & 24.0          & 9.4          & 5.8          & 8.6          & 20.6          \\
                                                                    & \textit{BADGR}           & FloorPlan-60K                 & \textbf{12.2} & \textbf{9.5} & \textbf{7.2} & \textbf{18.5} & \textbf{7.1} & \textbf{4.5} & \textbf{6.7} & \textbf{15.3} \\ \hline
\multirow{4}{*}{1}                                                  & \textit{CovisPose+}      & ZInD                          & 19.7          & 15.6         & 11.8         & 33.7          & 12.3         & 7.0          & 10.8         & 27.4          \\
                                                                    & \textit{BA Only}         & N/A                           & 17.6          & 12.6         & 12.0         & 34.2          & 12.6         & 6.7          & 11.4         & 27.5          \\
                                                                    & \textit{BADGR}           & RPLAN                         & 15.1          & 11.1         & 8.5          & 26.5          & 9.2          & 5.9          & 8.8          & 20.6          \\
                                                                    & \textit{BADGR}           & FloorPlan-60K                 & \textbf{11.2} & \textbf{8.8} & \textbf{6.5} & \textbf{18.6} & \textbf{7.0} & \textbf{4.6} & \textbf{6.6} & \textbf{15.8} \\ \hline
\multirow{4}{*}{2}                                                  & \textit{CovisPose+}      & ZInD                          & 17.9          & 14.5         & 10.2         & 30.8          & 12.0         & 7.0          & 9.0          & 26.0          \\
                                                                    & \textit{BA Only}         & N/A                           & 14.3          & 10.8         & 9.2          & 30.1          & 10.5         & 6.2          & 9.9          & 23.3          \\
                                                                    & \textit{BADGR}           & RPLAN                         & 14.1          & 10.5         & 8.6          & 27.8          & 9.1          & 5.6          & 8.7          & 20.1          \\
                                                                    & \textit{BADGR}           & FloorPlan-60K                 & \textbf{10.7} & \textbf{8.9} & \textbf{6.0} & \textbf{17.1} & \textbf{6.4} & \textbf{4.4} & \textbf{5.9} & \textbf{13.9} \\ \hline
\multirow{4}{*}{3}                                                  & \textit{CovisPose+}      & ZInD                          & 18.1          & 14.8         & 10.5         & 31.6          & 12.3         & 7.2          & 10.2         & 26.9          \\
                                                                    & \textit{BA Only}         & N/A                           & 13.4          & 10.7         & 8.2          & 30.5          & 10.6         & 6.2          & 9.0          & 22.1          \\
                                                                    & \textit{BADGR}           & RPLAN                         & 13.3          & 10.2         & 7.7          & 22.9          & 9.4          & 5.9          & 8.7          & 21.1          \\
                                                                    & \textit{BADGR}           & FloorPlan-60K                 & \textbf{10.6} & \textbf{8.9} & \textbf{6.0} & \textbf{17.5} & \textbf{6.6} & \textbf{4.3} & \textbf{6.0} & \textbf{14.6} \\ \hline
\end{tabular}
}
\end{table}

\section{Reprojection Errors} 

Reprojection errors are reported in Table \ref{tab:reproj_errs} to measure the view-consistency between the floor boundary projected from the predicted layout and poses and the per-image estimations, similar to the blue and green lines in the bottom-right images of Figure 1 of the main paper. The stats are computed from the per-column reprojection errors across all wall-assigned image columns, which is defined in Algorithm 1 of the main paper.

\begin{table}[H]
\centering
\vspace{-8pt}
\caption{Reprojection errors (\textit{L1} distance by pixel relative to an image size of 256 $\times$ 512) for wall-assigned columns, which measures view-consistency compared to the predicted floor boundary. Alongside Table 1 of the main paper, we observe that while \textit{BADGR} sometimes produces higher re-projection errors than \textit{BA-Only}, it consistently achieves lower layout and pose errors. This suggests that reprojection error influences accuracy but is not the sole factor in achieving high reconstruction accuracy. The stats are collected similarly to those from Table 1 of the main paper.}
\vspace{-6pt}
% Mean re-projection errors (by pixel) are reported for wall-assigned columns, measuring view consistency with the predicted floor boundary. Alongside Table \ref{tab: combined_zind}, we see that while \textit{BADGR} sometimes produces higher re-projection errors than \textit{BA}, it consistently achieves lower layout and pose errors, suggesting that re-projection error impacts accuracy but is not the sole factor for high reconstruction accuracy.

% BA can show a lower per-column reproj erro, but at a price of a lower global layout and poses. Additionally, because of GT is human labeled annotations that's not perfect, and inperfect noisy depth prediction. Explain refer to table 1. Use example. What is this GT? Generated from manually derived annotation. GT doesn't have depth. Reprojection error doesn't give the full story. See supplementaty for details.} 
\label{tab:reproj_errs}
\scalebox{0.8}{
\begin{tabular}{c|cccc|cccc|cccc|cccc}
\hline
Img/Rm                  & \multicolumn{4}{c|}{0.6}                                                                                            & \multicolumn{4}{c|}{1}                                                                                                       & \multicolumn{4}{c|}{2}                                                                                                       & \multicolumn{4}{c}{3}                                                                                                                \\
Method                  & Mn                       & Med                      & Std                               & p90                       & Mn                       & Med                      & Std                               & p90                                & Mn                       & Med                      & Std                               & p90                                & Mn                                & Med                      & Std                               & p90                               \\ \hline
\textit{CovisPose+}     & 1.38                     & 0.92                     & 1.81                              & 2.77                      & 1.52                     & 0.95                     & 2.08                              & 3.05                               & 1.69                     & 1.03                     & 2.40                              & 3.67                               & 1.79                              & 1.05                     & 2.57                              & 3.93                              \\
\textit{BA-Only}        & 0.70                     & \textbf{0.29}            & 1.21                              & 1.64                      & 0.78                     & \textbf{0.32}            & 1.46                              & 1.88                               & 0.81                     & 0.39                     & 1.48                              & 2.07                               & 0.89                              & 0.39                     & 1.45                              & 2.06                              \\
\textit{\textbf{BADGR}} & \textbf{0.65}            & 0.31                     & 1.17                              & \textbf{1.36}             & \textbf{0.75}            & \textbf{0.32}            & 1.34                              & 1.77                               & \textbf{0.78}            & \textbf{0.35}            & 1.37                              & 1.81                               & 0.81                              & \textbf{0.36}            & 1.33                              & 2.04                              \\
GT Scene + Predicted Boundary                      & \multicolumn{1}{l}{0.91} & \multicolumn{1}{l}{0.82} & \multicolumn{1}{l}{\textbf{0.56}} & \multicolumn{1}{l|}{2.77} & \multicolumn{1}{l}{0.90} & \multicolumn{1}{l}{0.80} & \multicolumn{1}{l}{\textbf{0.56}} & \multicolumn{1}{l|}{\textbf{1.55}} & \multicolumn{1}{l}{0.89} & \multicolumn{1}{l}{0.78} & \multicolumn{1}{l}{\textbf{0.56}} & \multicolumn{1}{l|}{\textbf{1.56}} & \multicolumn{1}{l}{\textbf{0.89}} & \multicolumn{1}{l}{0.77} & \multicolumn{1}{l}{\textbf{0.57}} & \multicolumn{1}{l}{\textbf{1.56}} \\ \hline
\end{tabular}
}
\vspace{-6pt}
\end{table}

As Table \ref{tab:reproj_errs} shows, overall reprojection error increases with the number of input images. This is caused by the accumulating pose errors and inconsistencies in floor boundary estimates across overlapping regions. Both \textit{BA-Only} and \textit{BADGR} consistently show lower reprojection errors compared to \textit{CovisPose+}. In most cases, \textit{BADGR} reports slightly lower reprojection errors than \textit{BA-Only},
likely because BA-Only can get stuck in local minima of the loss function and uses a PyTorch implementation that also considers memory and speed. In this implementation, adjustments are computed at per-column level and averaged to update poses and walls, rather than optimizing the total reprojection error across all columns.

\section{Coarse Scene Initialization for Inference}

\begin{figure}[h]
\begin{center}
\vspace{-4pt}
    \includegraphics[width=0.83\linewidth]{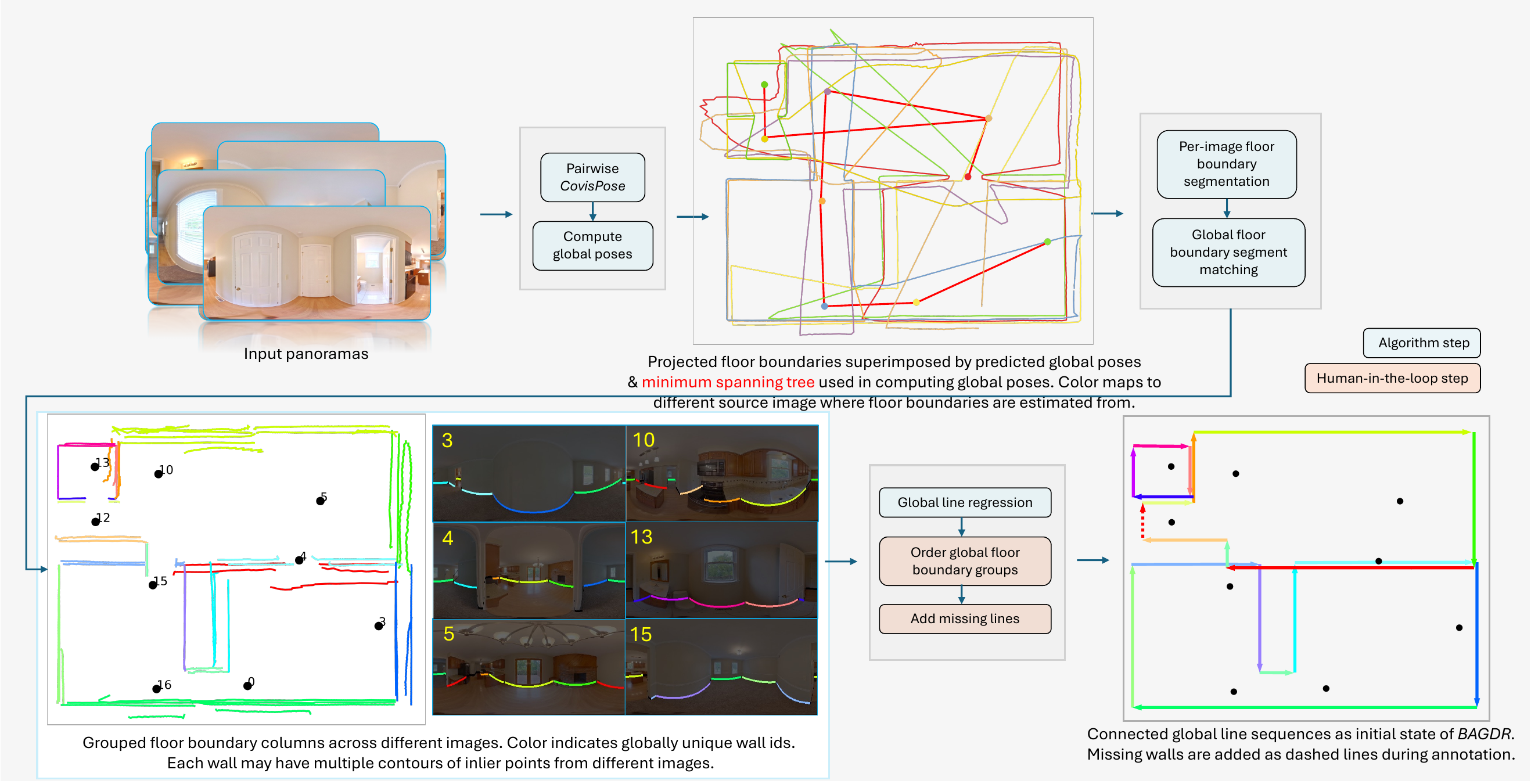}
    \vspace{-6pt}
    \caption{Overview of coarse scene initialization. }
    \label{fig:init}
\end{center}
\vspace{-12pt}
\end{figure}

\noindent\textbf{Initial Poses} From input panoramas $\{P^i \}$,  a modified \textit{CovisPose} model \cite{hutchcroft2022covispose} is executed exhaustively on each pair of panoramas from the same floor. 
This model has the same exact architecture as the original \textit{CovisPose} model predicting: 1) relative camera pose $\Tilde{E}^{(i,j)}\in SE(2)$, 2) floor boundaries $\{\Tilde{\mathcal{B}}^i \}$, 3) cross-view co-visibility, angular correspondences $\{\Tilde{\alpha}^{i,j} \}$, $\{\Tilde{\varphi}^{i,j} \}$. It additionally predicts binary classification of room corners $\{\Tilde{\mathcal{V}}^i \}$ for each column. The model is trained on the \textit{ZInD} dataset \cite{cruz2021zillow} with the same image pairs as \cite{hutchcroft2022covispose} and an additional corner loss function similar to that of \cite{sun2019horizonnet}. Pose pairs of co-visibility score greater than 0.1 are selected to create a minimal spanning tree of the pose graph using a greedy algorithm, similar to \cite{nejatishahidin2023graph}. Prior to computing global poses $\Tilde{E}^i$,  $\Tilde{E}^{i,j}$ are corrected through axis alignment with a 45° interval using predicted vanishing angles \cite{zhang2014panocontext}. 

\noindent\textbf{Initial Walls} The per-panorama floor boundary $\{\hat{\mathcal{B}}^i \}$ is segmented with room corners $\{\hat{\mathcal{V}}^i \}$. Inlier boundary points are then extracted with \textit{RANSAC}, and initial wall parameters \resizebox{0.09\hsize}{!}{$(\overrightarrow{v_{m,k}^i}$, $b_{m,k}^i)$} are computed for each local wall detected from panoramas ${P^i}$. Voting-based heuristics are used to match inlier boundary points, which maps to per-panorama local line segments, between panorama pairs using $\{\Tilde{\alpha}^{i,j} \}$, $\{\Tilde{\varphi}^{i,j} \}$ and {$(\overrightarrow{v_{m,k}^i}$, $b_{m,k}^i)$}. Pairwise local line matches are aggregated into a unique global wall identity for wall $l_{m,k}$ shared across ${P^i}$. The estimated wall parameters, i.e. {$(\overrightarrow{v_{m,k}}, \hat{b}_{m,k})$}, are computed with linear regression, where \resizebox{0.035\hsize}{!}{$\overrightarrow{v_{m,k}}$} is selectively axis-aligned with a $45^\circ$ interval. Only wall angles closer to $10^\circ$ to the vanishing directions, e.g. 0, 45, 90, 135, are corrected. Finally, an annotator uses a graphics interface to: 1) provide global wall connectivity (shown as arrows in the bottom right image of Figure \ref{fig:init}), and 2) add missing room corners with their rough initial positions with guidance from the images and topdown projected floor boundaries (dotted lines in the bottom right image of Figure \ref{fig:init}).  The number of room corners and wall orientations are static input to \textit{BADGR}. 

During testing, a subset of panoramas are selected as described in Section 8.1 of the main paper and Section \ref{sec:zillow_train} of the Supplement.
To generate the coarse initial layouts, we use the connectivity of the annotated global scene as discussed above, re-compute parameter {$(\overrightarrow{v_{m,k}^i}$, $b_{m,k}^i)$} of visible walls using the inlier boundary point from the selected panoramas, and inherit the parameters of invisible walls from the initial coarse scene generated with all available panoramas from the ZInD dataset. Only rooms with visible walls are included in the coarse initial layouts. \textit{PolyDiffuse} \cite{chen2024polydiffuse} also uses simple annotation during initialization. Our paper focuses on the difficult step of global refinement. Automating this annotation is future work to automate an end-to-end pipeline. 

\section{Discussion}
% \textbf{How is \textit{BADGR} compared to \textit{PuzzleFusion} \cite{hossieni2024puzzlefusion} and \textit{Extreme SfM}\cite{shabani2021extreme}, as both approaches can solve floor plan layouts and camera poses?} 
\textbf{\textit{PuzzleFusion}} (\textit{PF}) \cite{hossieni2024puzzlefusion} and \textbf{\textit{Extreme SfM}} (\textit{E-SfM}) \cite{shabani2021extreme} also produce floor plan layouts and camera poses. Here we provide a discussion on their differences to \textit{BADGR}. Both \textit{PF} and \textit{E-SfM} estimate the rotation and translation of given unposed non-deformable room layouts by solving jigsaw puzzles. 
Camera poses are then inferred from the puzzle solution. This has different objectives than ours: 1) within the same room their relative positions among individual walls and multiple camera poses stay unchanged; 2) neither method uses information from a set of horizontal-facing images without precise poses as input constraints to guide optimization for view-consistency. Both can be used for initialization of \textit{BADGR} like \textit{CovisPose}.
Code and weights of \textit{PF} trained on RPLAN aren't publicly available. We contacted the authors, and the code no longer runs. \textit{E-SfM} takes hours or even days to process a single house \cite{hossieni2024puzzlefusion}, so neither can be used as baselines. \textit{BADGR} solves a different task as we are deforming room shapes. Instead, we simulate \textit{BADGR} refining \textit{PF}-initialized layouts by adding Gaussian noise (10.55 \textit{Mean Positional Error in pixels} (MPE) matches \textit{PF}) to room translations, with relative poses among cameras and walls within each room given for initialization. \textit{BADGR} reduces the MPE$\downarrow$ of room placement from 10.55 or 4.1\% (normalized by $256\times256$ pixel resolution) (full RPLAN) to 0.93\% (77.3\% lower), calculated by average shift of each vertex. We also report 0.98\%, 1.45\% mean translation errors in layout and poses. MPE of \textit{E-SfM} is only reported for small RPLAN as 29.44 or 11.5\% \cite{hossieni2024puzzlefusion}.

\textbf{\textit{GraphCovis}} \cite{nejatishahidin2023graph} and \textbf{\textit{PanoPose}} \cite{tu2024panopose} didn't publicly release their code. \textit{GraphCovis} estimates global poses among up-to-5 input panorama images. We compared the pose errors of \textit{BADGR} with \textit{GraphCovis} under the similar input settings originally evaluated on ZInD \cite{cruz2021zillow} in table \ref{tab:graphcovis}. It demonstrates \textit{BADGR}'s robust performance among different sizes of homes and missing room scenarios.

\begin{table}[H]
\centering
\vspace{-6pt}
\caption{Statistics of \textit{absolute translation error} and \textit{absolute rotation error} on group of three, four, and five panoramas for \textit{GraphCovis} (\textit{GC}) and \textit{BADGR}. The accuracy of \textit{GraphCovis} is imported from Table 1 of \cite{nejatishahidin2023graph}.}
\label{tab:graphcovis}
\vspace{-6pt}
\scalebox{0.7}{
\begin{tabular}{ccccccccccccc}
\hline
\multirow{2}{*}{\begin{tabular}[c]{@{}c@{}}Group-Size\\ \# imgs\end{tabular}} & \multicolumn{3}{c}{\textit{GC} Rot $\downarrow$ ($^\circ$)} & \multicolumn{3}{c}{\textit{BADGR} Rot $\downarrow$ ($^\circ$)}           & \multicolumn{3}{c}{\textit{GC} Transl. $\downarrow$ (cm)} & \multicolumn{3}{c}{\textit{BADGR} Transl. $\downarrow$ (cm)}      \\ \cline{2-13} 
                            & Mn        & Med      & Std       & Mn            & Med           & Std           & Mn            & Med          & Std  & Dist (cm)     & Med          & Std          \\ \hline
3                           & 2.00      & 0.85     & 9.15      & \textbf{0.25} & \textbf{0.20} & \textbf{0.28} & \textbf{8.1}  & \textbf{3.8} & 29.2 & 9.2           & 6.1          & \textbf{5.9} \\
4                           & 3.19      & 0.94     & 13.36     & \textbf{0.26} & \textbf{0.22} & \textbf{0.30} & 15.3          & \textbf{6.1} & 43.0 & \textbf{10.6} & \textbf{6.1} & \textbf{6.0} \\
5                           & 3.29      & 1.07     & 12.04     & \textbf{0.25} & \textbf{0.19} & \textbf{0.30} & 17.2          & 8.2          & 38.4 & \textbf{10.4} & \textbf{6.3} & \textbf{6.7} \\ \hline
\end{tabular}
}
\vspace{-12pt}
\end{table}

\section{Failure Cases}

\begin{figure}[ht]
\vspace{-6pt}
\begin{center}
    \includegraphics[width=0.95\linewidth]{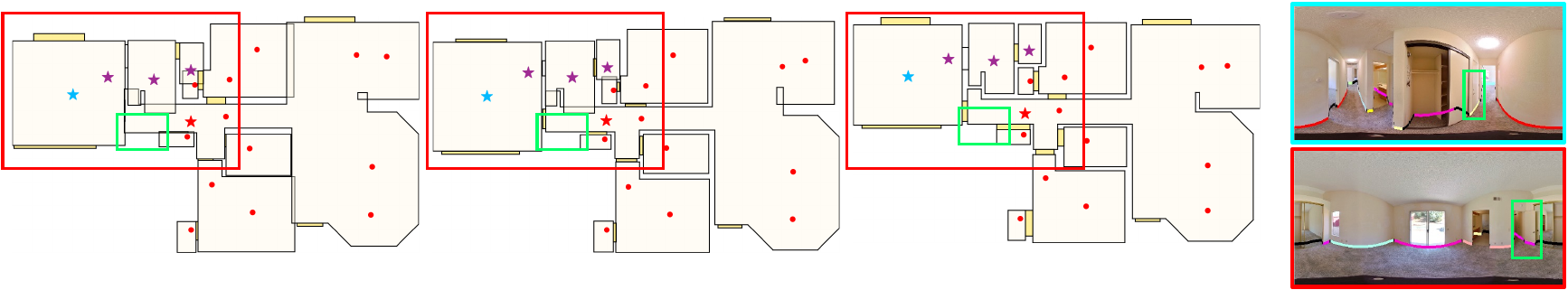}
    \vspace{-8pt}
    \caption{Failure case example caused by errors in coarse scene initialization. The colored lines in the images on the right represent estimated floor boundaries, with colors indicating their assigned unique global wall id. The heuristic failed to match two wall segments (highlighted in \textcolor{green}{rectangles}) using dense column correspondences and floor boundaries from the \textit{}{CovisPose} model. Additionally, the large initial error in the highlighted section (highlighted in \textcolor{red}{rectangles}) may fall near the boundary of the 20-step truncated diffusion data distribution, contributing to the issue.}
    \vspace{-18pt}
    \label{fig:failure_init}
\end{center}
\end{figure}

We present three failure cases to highlight the challenges and opportunities for \textit{BADGR}. Overall, \textit{BADGR} achieves high accuracy when input images are minimally-connected by covisible walls through column-to-wall assignments from the initial coarse scene. However, since \textit{BADGR} is trained on simulated panorama poses and column-to-wall assignments, the model can struggle when faced with scenarios outside the training distribution. An example is shown in Figure \ref{fig:failure_init}, where the initial scene contains large errors over wide areas, and the column-to-wall assignments fail to establish critical covisible walls between panoramas. This underscores the need for future development of an end-to-end initialization method to establish global column-to-wall assignments.

\begin{figure}[h]
\vspace{-12pt}
\begin{center}
    \includegraphics[width=0.95\linewidth]{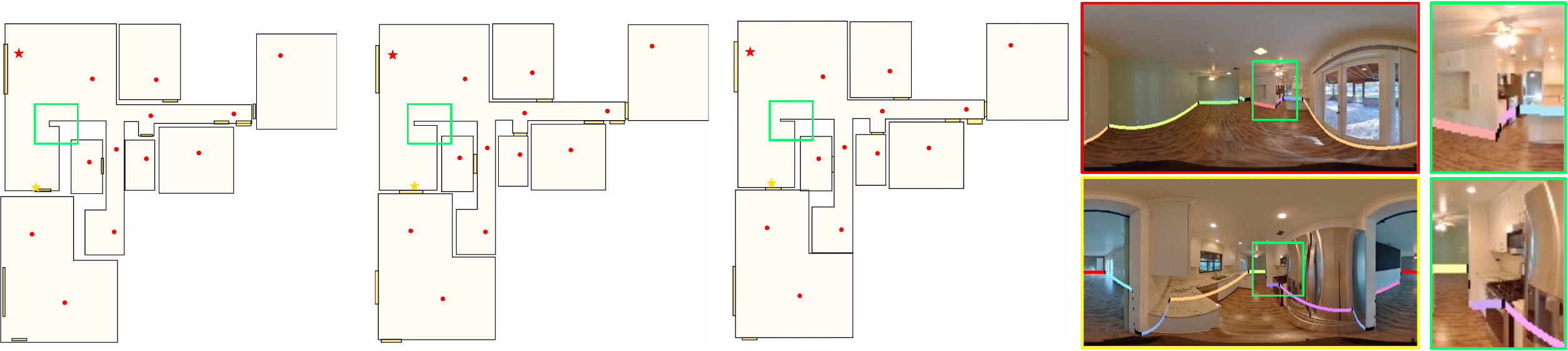}
    \caption{Failure case example where the \textcolor{red}{highlighted} wall is predicted with an incorrect length due to the limited image column coverage. The colored lines in image views represent similarly as in Figure \ref{fig:failure_init}.}
    \vspace{-16pt}
    \label{fig:failure_no_corner}
\end{center}
\end{figure}

\textit{BADGR} relies on floor boundaries for positional information along the normal direction of the target surface. This explains the failure case in Figure \ref{fig:failure_no_corner}, where the wall length is estimated incorrectly due to a lack of guidance to the model. Future work could incorporate cues from wall junctions, similar to \cite{sun2019horizonnet, wang2021led2}, or encode pixel positions of pre-assigned columns to better constrain visible walls and infer invisible wall positions. 

\begin{figure}[h]
\vspace{-8pt}
\begin{center}
    \includegraphics[width=0.94\linewidth]{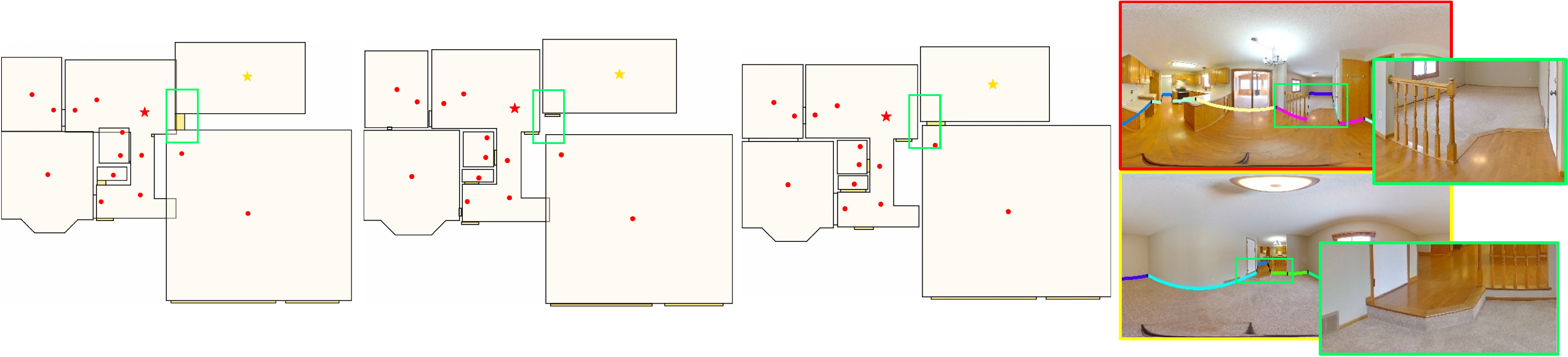}
    \vspace{-6pt}
    \caption{Failure case example due to inconsistent floor heights. The colored lines in image views represent similarly as in Figure \ref{fig:failure_init}.}
    \label{fig:failure_height}
    \vspace{-18pt}
\end{center}
\end{figure}

\textit{BADGR} assumes consistent floor heights throughout the area. When this assumption is violated, such as with a sunken floor (Figure \ref{fig:failure_height}), planar \textit{BA} may place walls farther than their actual positions. Future work may include extending \textit{BADGR} to represent varying camera height and wall heights, and expanding the training data to cover this issue.

\section{More Qualitative Results}

The reprojected wall lines in the image are drawn with a thickness equal to 1.1\% of the image height. This thickness can cover significant floor distances, especially when walls are near the image center. For example, at a pitch angle of 30° (floor distance of 0.58 camera heights), the line covers 4.5\% of the camera height; at 45° (1 camera height), it covers 7.4\%; at 60° (1.73 camera height), 14.7\%; and at 75° (3.73 camera height), 55.2\%. While the blue and green lines represented in small images sometimes appear to overlap, particularly for walls farther from the camera, \textit{BADGR} processes continuous inputs and outputs for coordinates, enabling higher precision. See quantitative results for more precise details.
% 0039 Blue yellow image, no visual overlap, only layout constraints.

\begin{figure}[t]
\begin{center}
    \includegraphics[width=0.96\linewidth]{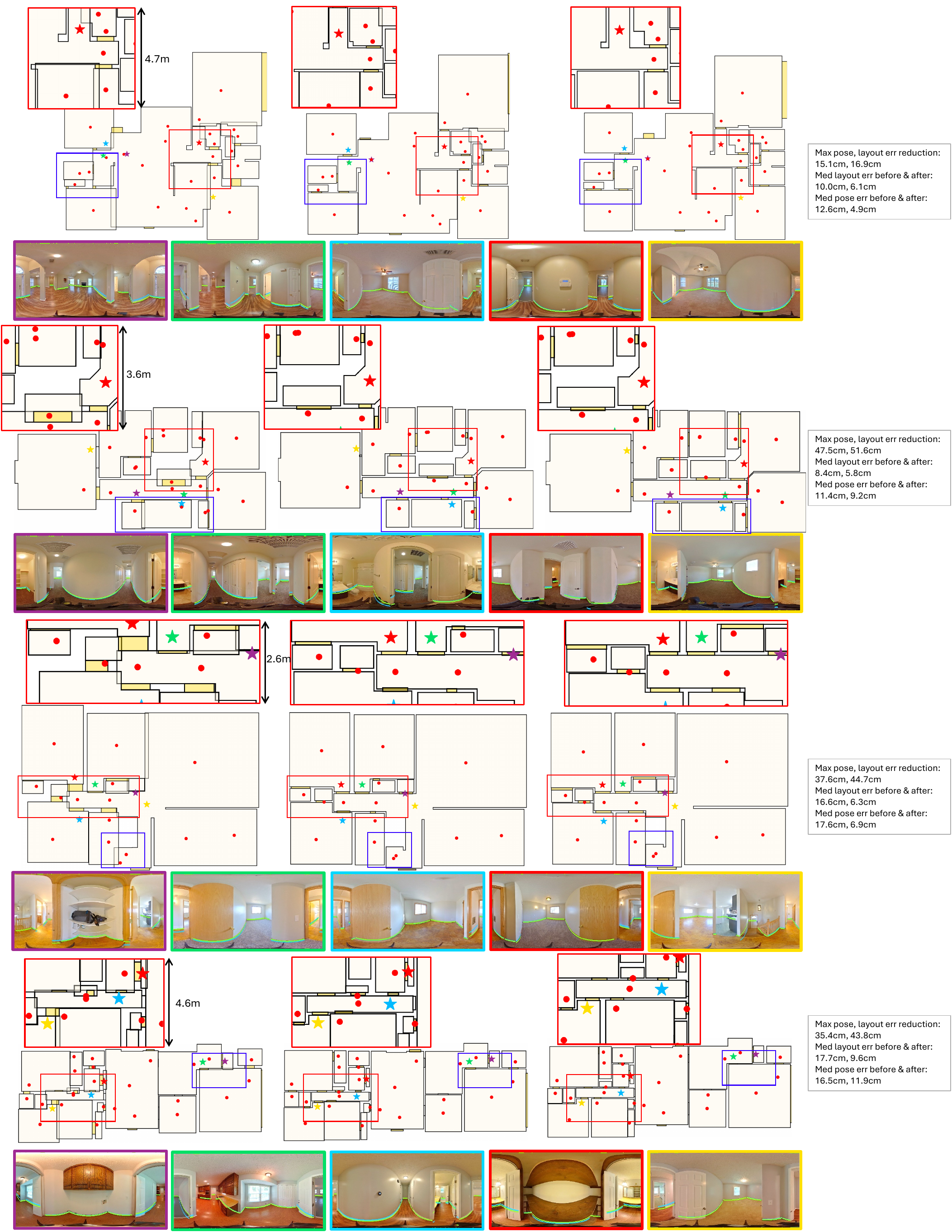}
    \caption{More qualitative results trained on FloorPlan-60K dataset and tested on ZInD dataset (page 1), with input densities at a maximum of 2 input images from each input partial room. The topdown views from left to right are before, after \textit{BADGR} optimization and the ground truth.}
    \label{fig:more_quals_p2}
\end{center}
\end{figure}

\begin{figure}[t]
\begin{center}
    \includegraphics[width=0.96\linewidth]{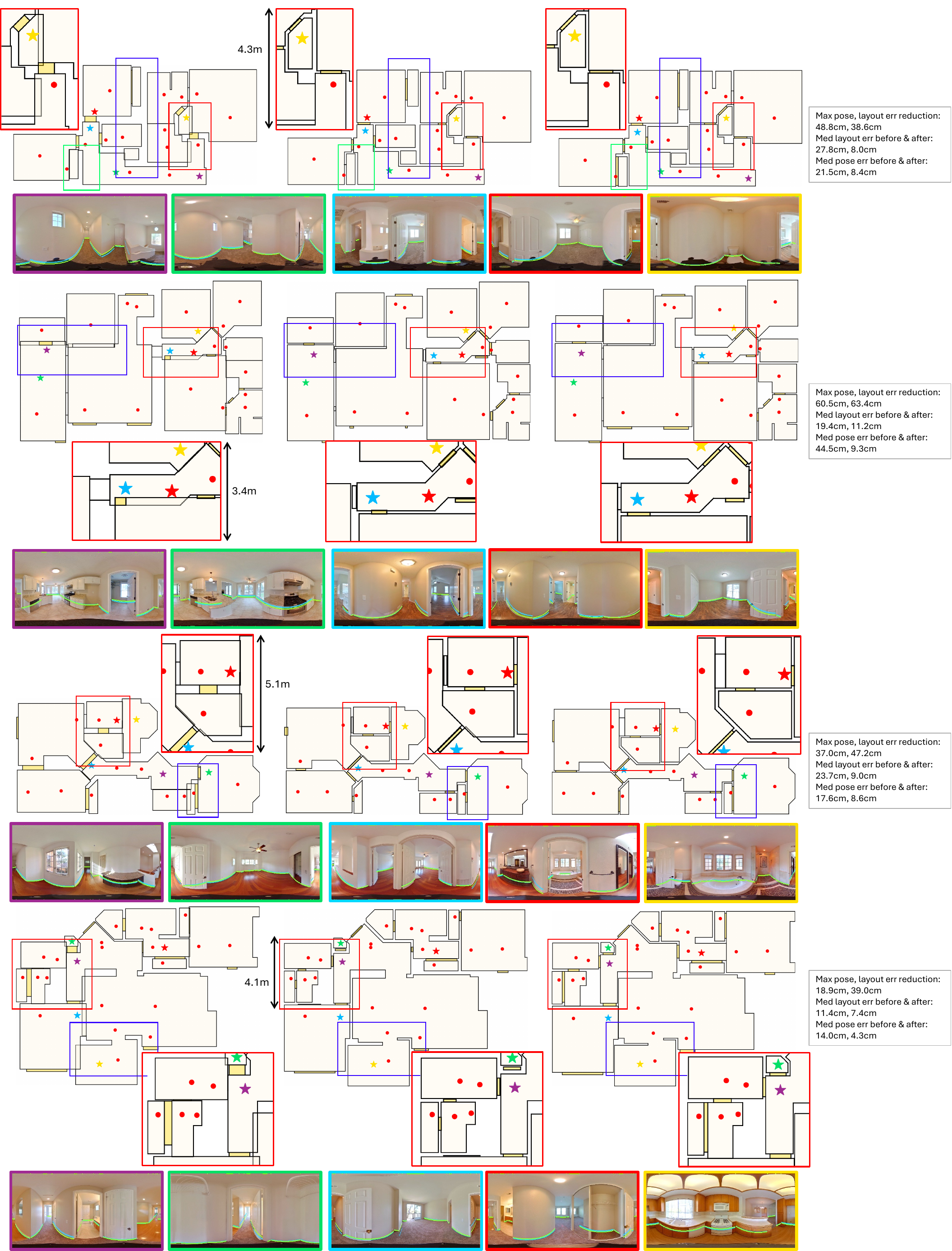}
    \caption{More qualitative results trained on FloorPlan-60K dataset and tested on ZInD dataset (page 2), with input densities at a maximum of 2 input images from each partial room. The topdown views from left to right are before, after \textit{BADGR} optimization and the ground truth.}
    \label{fig:more_quals_p3}
\end{center}
\end{figure}

\begin{figure}[t]
\begin{center}
    \includegraphics[width=0.96\linewidth]{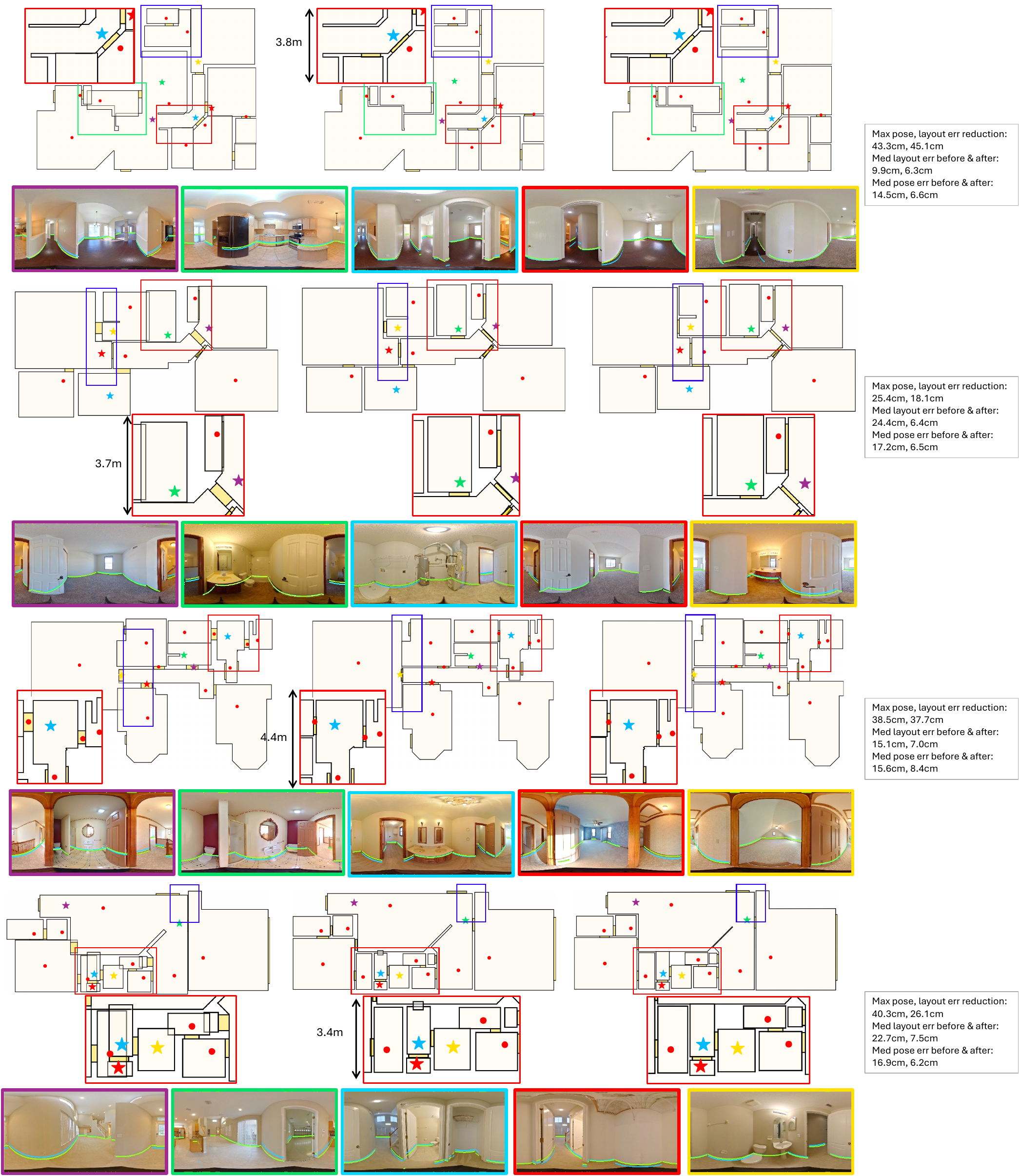}
    \caption{More qualitative results trained on FloorPlan-60K dataset and tested on ZInD dataset (page 3), with input densities at a maximum of 1 input images from each partial room. The topdown views from left to right are before, after \textit{BADGR} optimization and the ground truth.}
    \label{fig:more_quals_p1}
\end{center}
\end{figure}

\begin{figure}[t]
\begin{center}
    \includegraphics[width=0.73\linewidth]{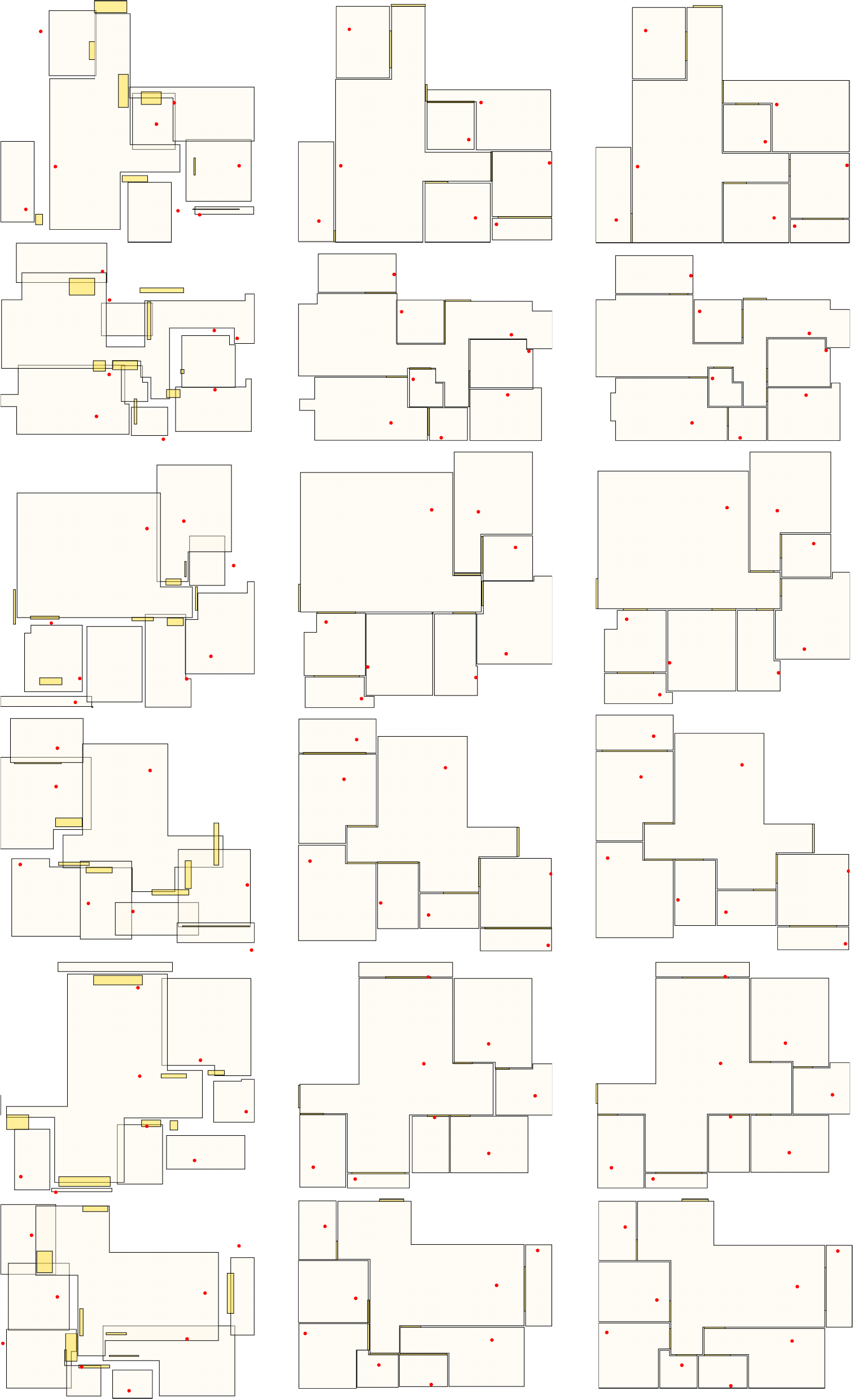}
    \caption{Qualitative results trained and tested on RPLAN dataset (page 4), with input densities at one input image from each partial room. The topdown views from left to right are before, after \textit{BADGR} optimization and the ground truth. The initial state is created by adding Gaussian noise from 20-step diffusion q-sampling \cite{ho2020ddpm} into the ground truth poses and layouts. Details see Section 8.2 of the main paper. This figure demonstrates \textit{BADGR}'s capability to refine initial scenes with much higher noise than those from the ZInD test cases. }
    \label{fig:more_quals_p4}
\end{center}
\end{figure}

\clearpage


\end{document}